\documentclass[accepted]{uai2026}

\usepackage[american]{babel}

\usepackage{amsmath,amssymb}
\usepackage{graphicx} 

\newcommand{\eg}{\textit{e.g.}}
\newcommand{\ie}{\textit{i.e.}}
\newcommand{\Var}{\mathrm{Var}}
\newcommand{\E}{\mathbb{E}}

\usepackage{xcolor}
\usepackage{tcolorbox}
\usepackage{amsthm}
\tcbuselibrary{theorems}
\definecolor{defback}{RGB}{248, 254, 252}
\definecolor{defframe}{RGB}{49, 122, 104}
\newtcolorbox{defbox}[1][]{colback=defback,colframe=defframe,boxsep=0pt,grow to left by=4pt,left=10pt,grow to right by=4pt,right=10pt,top=10pt,bottom=10pt,#1}
\definecolor{theback}{RGB}{235, 250, 255}
\definecolor{theframe}{RGB}{30, 100, 160}
\newtcolorbox{thmbox}[1][]{colback=theback,colframe=theframe,boxsep=0pt,grow to left by=4pt,left=10pt,grow to right by=4pt,right=10pt,top=10pt,bottom=10pt,#1}
\definecolor{corback}{RGB}{246, 245, 252}
\definecolor{corframe}{RGB}{41, 32, 116}
\newtcolorbox{corbox}[1][]{colback=corback,colframe=corframe,boxsep=0pt,grow to left by=4pt,left=10pt,grow to right by=4pt,right=10pt,top=10pt,bottom=10pt,#1}

\definecolor{colorf}{HTML}{FF9396}
\definecolor{colors}{HTML}{FFC991}
\definecolor{colort}{HTML}{FFF6A9}

\newtheorem{theorem}{Theorem}

\theoremstyle{remark}

\theoremstyle{plain}
\newtheorem{corollary}{Corollary}

\theoremstyle{definition}
\newtheorem{definition}{Definition}

\usepackage{booktabs}
\usepackage{multirow}

\usepackage{pifont}
\newcommand{\yes}{\textcolor{black}{\ding{51}}}
\newcommand{\no}{\textcolor{black}{\ding{55}}}
\colorlet{ignore}{black!50!white}

\usepackage{natbib} 
\usepackage{mathtools} 
\usepackage{booktabs} 
\usepackage{tikz} 

\title{Score-Regularized Joint Sampling with Importance Weights for Flow Matching}

\author{
Xinshuang Liu \quad Runfa Blark Li \quad Shaoxiu Wei \quad Truong Nguyen \\
University of California, San Diego \\
San Diego, CA, USA \\
{xil235@ucsd.edu \quad rul002@ucsd.edu \quad shwei@ucsd.edu \quad tqn001@ucsd.edu} \\
\url{https://XinshuangL.github.io/SRIW-Flow}
}

\begin{document}
\maketitle

\begin{abstract}
Flow matching models effectively represent complex distributions, yet estimating expectations of functions of their outputs remains challenging under limited sampling budgets. Independent sampling often yields high-variance estimates, especially when rare but high-impact outcomes dominate the expectation. We propose a \textbf{non-IID sampling framework} that jointly draws multiple samples to cover diverse, salient regions of a flow matching model's generative distribution. To balance diversity and quality, we introduce a \textbf{score-based regularization for the diversity mechanism} (SR), which uses the score function, \textit{i.e.}, the gradient of the log probability, to ensure samples are pushed apart within high-density regions of the data manifold, mitigating off-manifold drift. To enable unbiased estimation when desired, we further develop an approach for \textbf{importance weighting of non-IID flow samples} by learning a residual velocity field that reproduces the marginal distribution of the non-IID samples and by evolving importance weights \emph{along trajectories}. Empirically, our method produces diverse, high-quality samples and accurate importance-weight estimates and debiased expectation estimates, advancing the reliable characterization of flow matching model outputs.
\end{abstract}

\section{Introduction}
\label{sec:intro}

Flow matching models~\citep{DBLP:conf/iclr/LipmanCBNL23} are powerful tools for representing complex distributions. Beyond drawing individual samples, many applications require expectations of \emph{functions} of model outputs~\citep{DBLP:journals/corr/abs-2502-03687}. Formally, for a fixed, pre-trained flow matching model with distribution $p(x)$ and a task-defined functional $f:\mathcal{X}\!\to\!\mathbb{R}^d$, we seek
\begin{equation}
\mu = \mathbb{E}_{X\sim p}\big[f(X)\big].
\end{equation}
As an illustrative example, consider estimating the expected rating of a text-to-image model under a fixed prompt, where $f(X)$ rates a generated image $X$ (\eg, a user rating) and $\mu$ is the model's expected score. A common practice is IID Monte Carlo with $X^{(1)},\dots,X^{(n)}\!\sim\!p$,
\begin{equation}
\hat{\mu}_{\text{IID}} = \frac{1}{n}\sum_{i=1}^n f\big(X^{(i)}\big).
\end{equation}
However, when sampling is costly, $n$ is small and $\hat{\mu}_{\text{IID}}$ can have high variance, especially for rare but high-impact outcomes (\eg, occasional surprising or disappointing images).
To cover such modes under a fixed budget, we draw the $n$ samples \emph{jointly} and encourage them to be diverse. Enforcing diversity, however, changes each sample's marginal distribution, so a plain average is biased; \emph{importance weighting} corrects this, recovering an unbiased estimate of $\mu$. This motivates pairing \emph{joint, diversity-enhancing sampling} with \emph{importance weighting}.

\begin{figure}[t]
    \centering
    \includegraphics[width=\linewidth]{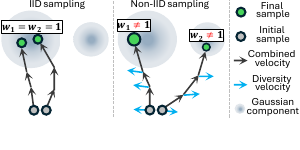}
    \caption{\textbf{Illustration of importance-weighted non-IID sampling.} Under IID sampling, both samples are likely drawn from the same dominant mode. In contrast, diversity velocity encourages samples to diverge along their trajectories, leading to coverage of multiple modes. To correct the resulting sampling bias, importance weights are required; intuitively, $w_1\!>\!1\!>\!w_2$, since non-IID sampling draws the second sample from a minor mode.}
    \label{fig:non_iid_and_importance}
\end{figure}

\paragraph{From joint sampling to unbiased estimation.}
Let $X^{(1:n)}\!\sim\!p_{\text{joint}}$ denote $n$ samples drawn \emph{once} from a joint distribution that encourages diversity. Denote by $p'(x)$ the marginal density of any single element under $p_{\text{joint}}$. An unbiased estimator is then (see Appendix~\ref{app:importance_sampling} for details)
\begin{equation}
\hat{\mu}_{\text{N-IID}} = \frac{1}{n}\sum_{i=1}^n w\big(X^{(i)}\big)\,f\big(X^{(i)}\big),
\quad
w(x)=\frac{p(x)}{p'(x)}.
\label{eq:importance_weight}
\end{equation}
Thus, a \emph{good} non-IID strategy for estimating $\mu$ must satisfy two goals: \textbf{(G1) Diversity with quality}---the jointly drawn set covers $p$'s support while keeping each sample on-manifold; and \textbf{(G2) Unbiasedness}---we can accurately estimate the per-sample importance weights $w(x)$.

\paragraph{Limitations of existing diverse samplers.}
Recent work explores non-IID sampling to improve diversity. \emph{Particle Guidance} (PG)~\citep{DBLP:conf/iclr/CorsoXBBJ24} couples diffusion trajectories via a repulsive potential, while \emph{DiverseFlow} (DF)~\citep{DBLP:conf/cvpr/MorshedB25} injects diversity into flow matching via a determinantal point process (DPP) objective~\citep{kulesza2012determinantal}. Both can be viewed as augmenting the sampling dynamics with a \emph{diversity velocity} that pushes concurrent trajectories apart. When jointly sampling $n$ trajectories $X^{(1:n)}_t$ from a flow matching model, these methods solve $n$ coupled ODEs
\begin{equation}
    \dot X_t^{(i)} = v \big(X_t^{(i)}, t\big) + u\big(X^{(i)}_t,X^{(-i)}_t,t\big), \!\quad X_0^{(i)} \overset{\text{IID}}{\sim} p_0,
    \label{eq:diversity_enhanced_ODEs}
\end{equation}
where $i$ indexes sample trajectories, $X^{(-i)}_t$ denotes the set excluding $X^{(i)}_t$, $t \in [0,1]$, $v$ is the pre-trained flow-matching velocity field (parameters omitted since they remain fixed), and $u$ computes a diversity velocity that pushes samples apart. Although effective at increasing diversity, these methods face a trade-off: strong diversity velocities improve spread but risk pushing samples into low-density, off-manifold regions; weak velocities preserve quality but limit diversity gains. Moreover, these approaches do \emph{not} provide the importance weights needed to form unbiased estimators of expectations like $\mu$, so equal-weighted averages generally introduce bias.

\paragraph{Overview of our approach.}
We propose a \emph{score-regularized non-IID sampling framework with importance weights} that achieves both (G1) and (G2) (Figure~\ref{fig:non_iid_and_importance}). For a pre-trained flow matching model $v$ and $n$ coupled trajectories $\{X_t^{(i)}\}_{i=1}^n$, we \emph{constrain the direction} of $g$ using the model's score $\nabla_x \log p(x\,|\,t)$ so diversity pushes samples \emph{within} high-density regions (along the data manifold) rather than ejecting them off-manifold. Since $g$ will be normalized in practice, removing components that push samples off-manifold will amplify on-manifold components. This yields sets that are both diverse and high-quality, alleviating the core trade-off observed in prior work.
To obtain unbiased estimates, we provide (to our knowledge) the \emph{first} method to compute importance weights for non-IID, jointly sampled outputs from flow-matching models. Directly estimating the marginal $p'(x)$ of the joint sampler is challenging because $X^{(1:n)}$ is drawn only once. We address this by learning a lightweight \emph{residual velocity} $r_\phi(x,t)$ such that the perturbed flow
\begin{equation}
\dot{X}_t = v(X_t,t) + r_\phi(X_t,t), \quad X_0 \sim p_0,
\end{equation}
matches the \emph{marginal} distribution induced by the diversity-coupled sampler at $t\!=\!1$. Since rectified-flow models~\citep{DBLP:conf/iclr/LiuG023} are widely used in recent flow matching work~\citep{DBLP:conf/icml/EsserKBEMSLLSBP24,labs2025flux1kontextflowmatching,flux2024}, we focus on rectified flows here and include discussions for general flow-matching models in Section~\ref{subsec:method_importance}. For rectified flows, learning such a residual velocity field is sufficient for importance-weight estimation, and a smaller network for $r_\phi$ than for $v$ can be used to reduce training and inference cost. Once trained for a given joint-sampling configuration, the residual can be reused across many test evaluations, so the training cost is amortized across the expectations and thus is negligible.

\paragraph{Empirical validation.}
We evaluate our method comprehensively. For accurate diagnosis, we first test on a Gaussian mixture model, where the true density (under IID sampling) is available in closed form. Our method improves sample diversity and quality and yields more accurate importance-weight estimates and debiases expectation estimation. We then evaluate advanced conditional flow-matching models: Stable Diffusion 3.5 Medium~\citep{DBLP:conf/icml/EsserKBEMSLLSBP24} for text-to-image generation and FLUX.1-Fill-dev~\citep{labs2025flux1kontextflowmatching,flux2024} for image inpainting, where our method improves coverage of the output distribution as a plug-in via score-based regularization. Across these tasks, our approach produces diverse, high-quality samples and accurate importance-weight estimates that debias expectation estimation, and demonstrates potential gains for large models.

\textbf{Contribution and Significance.}
We propose a non-IID sampling framework for flow matching. We constrain diversity directions using model scores to maintain sample quality while enhancing coverage. For unbiased expectation estimation, we further introduce a novel method to compute importance weights for non-IID flow samples. The correctness of the proposed method is theoretically proven, followed by comprehensive empirical validation of its effectiveness. By open-sourcing our code, this work facilitates the management of the diversity--quality trade-off of flow matching generative models and effective importance weight estimation for unbiased expectation estimation, all of which are fundamental and important in flow matching generative models.

\section{Preliminaries \& Related Work}

\subsection{Flow Matching Models}

Flow matching models~\citep{DBLP:conf/iclr/LipmanCBNL23} generate a sample $X_1$ by integrating an ODE from a base distribution to the target distribution
\begin{equation}
    \dot{X}_t = v_\theta(X_t, t), \quad X_0 \sim p_0\,,
\end{equation}
from $t=0$ to $t=1$, where $v_\theta(x,t)$ is a learned time-dependent velocity field and $p_0$ is a tractable base density (\textit{e.g.}, a standard Gaussian). The ODE can be solved by a first-order integrator such as Euler's method:
\begin{equation}
    X_{t_{i+1}} = X_{t_i} + (t_{i+1} - t_i)\, v_\theta(X_{t_i}, t_i),
\end{equation}
where $0\!=\!t_0\!<\!\cdots\!<\!t_N\!=\!1$ and $N$ is the number of integration steps. The evolution of the sample density $p_t(x)$ at a fixed $x$ is given by the continuity equation:
\begin{equation}
\partial_t p_t(x) 
=-\nabla_x \cdot \big(v_\theta(x,t) p_t(x) \big), 
\! \quad p_{t=0}(x) = p_0(x).
\label{eq:continuity}
\end{equation}

One widely used variant of flow matching is the \emph{rectified flow} model~\citep{DBLP:conf/iclr/LiuG023}, adopted by Stable Diffusion 3~\citep{DBLP:conf/icml/EsserKBEMSLLSBP24} due to its effectiveness.
In rectified flow, given a sample $X_0 \sim p_0$ and a target sample $X_1$ (data), the intermediate state is defined by linear interpolation $X_t = (1-t)\,X_0 + t\,X_1$. The velocity field is then trained by 
\begin{equation}
    \min_\theta \mathbb{E}_{X_0, X_1,t}\Big[\big\|(X_1 - X_0) - v_\theta(X_t, t)\big\|^2_2\Big],
\end{equation}
which leads to $v_\theta(x,t) \approx \mathbb{E}[\,X_1 - X_0 \mid X_t = x\,]$, meaning the learned velocity at any point $x$ approximates the expected remaining change needed to reach the target. An important consequence is that the model's score function $s(x,t)$ can be computed in closed form from the velocity field and the current sample (Appendix~\ref{app:score_function})
\begin{equation}
    s(x,t) := \nabla_x \log p_\theta(x\,|\,t) \approx \frac{t\, v_\theta(x,t) - x}{1 - t}.
    \label{eq:score_velocity}
\end{equation}
When we use a trained velocity field, this identity is an approximation. We will later use this relationship to guide diversity in our sampling method. 

\subsection{Enhancing Diversity of Flow Matching Models}
\label{subsec:preliminary_diversity}

When drawing multiple samples concurrently from a generative model, a key goal is to ensure they cover different high-probability regions of $p_\theta$---in other words, to encourage \emph{diversity}. Recent methods for diffusion~\citep{Rombach_2022_CVPR} and flow~\citep{DBLP:conf/iclr/LipmanCBNL23} models have explored non-IID sampling strategies to increase diversity among $n$ joint samples~\citep{DBLP:conf/iclr/CorsoXBBJ24, DBLP:conf/cvpr/MorshedB25,liu2026consistency}. The core idea is to introduce a \emph{diversity objective} $h(X^{(1:n)}_t)$, where higher $h$ indicates a more diverse set.

A common construction is to define $h$ using normalized pairwise distances $K$ between samples:
\begin{equation}
K_{ij} = \frac{D_{ij}}{\text{med}(D)}, \quad D_{ij} = \big\|x^{(i)}_t - x^{(j)}_t \big\|^2_2,
\label{eq:K_definition}
\end{equation}
where $D_{ij}$ is the squared distance between samples $i$ and $j$ at time $t$, and $\text{med}(D)$ is the median of all $D_{ij}$ with $i\neq j$ (treated as a fixed constant to avoid affecting gradients). In practice, the diversity term can be computed using the predicted final samples $\hat x_1^{(i)} = x_t^{(i)}+v(x_t^{(i)},t)(1-t)$~\citep{DBLP:conf/cvpr/MorshedB25}. We adopt this formulation in all experiments; we omit it from analytical discussion, as it does not affect the conclusions.

To increase $h$ during sampling dynamics, one can compute the gradient of $h$ with respect to each sample as
\begin{equation}
    g\big(X^{(i)}_t, X^{(-i)}_t, t\big) = \nabla_{X^{(i)}_t}\, h\big(X^{(1:n)}_t\big),
\end{equation}
and use a scaled version as the \emph{diversity velocity} $u$ (as in Eq.~(\ref{eq:diversity_enhanced_ODEs})):
\begin{equation}
    u\big(X^{(i)}_t, X^{(-i)}_t, t\big) = \gamma\big(X^{(1:n)}_t, t\big) g\big(X^{(i)}_t, X^{(-i)}_t, t\big). 
\end{equation}
We set 
\begin{equation}
\gamma\big(X^{(1:n)}_t, t\big) = \lambda \sqrt{\,1-t\,}\;\frac{\Big\|\!v(X^{(1:n)}_t, t)\!\Big\|_2}{\Big\|\,\nabla_{X^{(1:n)}_t} h\!\big(X^{(1:n)}_t\big)\,\Big\|_2},
\label{eq:diversity_velocity}
\end{equation}
where $v(X^{(1:n)}_t, t)$ is the concatenation of velocities for all $X^{(i)}_t$ and $\lambda$ controls the intensity. As $t{\to}1$, the factor $\sqrt{\,1-t\,}$ shrinks the diversity velocity to preserve quality, and the ratio of norms keeps $u$ comparable in magnitude to $v$. \emph{This augmentation spreads the samples but faces a trade-off}: if $u$ is too strong, samples may be pushed off-manifold (quality drops); if too weak, diversity gains are limited. Moreover, \emph{existing approaches do not provide importance weights for jointly drawn samples; equal-weighted averages generally yield biased estimates.}

\section{Methodology}
\label{sec:method}

Our approach addresses both (G1) \emph{diversity with quality} and (G2) \emph{unbiasedness} by modifying the sampling of a pre-trained flow. We generate a batch of $n$ samples jointly, $X^{(1:n)} = (X^{(1)}, \dots, X^{(n)})$, from velocity $v(x,t)$ with base distribution $p_0(x)$. We encourage diversity while preserving on-manifold quality and assign an importance weight to each sample so that the expectation estimator remains unbiased (as in Eq.~(\ref{eq:importance_weight})). We achieve this with two components: (1) \emph{score-based diversity-velocity regularization} that pushes samples apart primarily along high-density directions, and (2) a \emph{residual velocity for importance weighting} added to $v$ to model each sample's marginal distribution under the joint sampler.

\begin{figure}[ht]
    \centering
    \includegraphics[width=\linewidth]{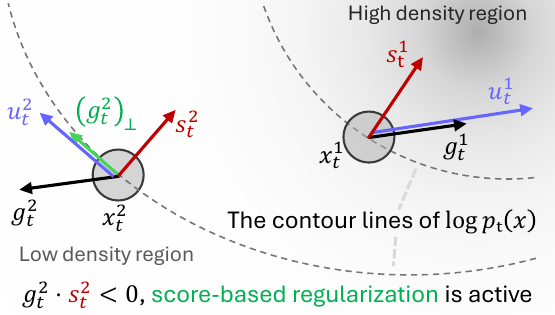}
    \caption{Illustration of our score-based regularization for the diversity mechanism.}
    \label{fig:diversity_force}
\end{figure}

\subsection{Score-based Diversity Velocity Regularization}
\label{subsec:diversity_velocity_regularization}

\paragraph{Gradient of a diversity term.}
As in Section~\ref{subsec:preliminary_diversity}, define a diversity objective $h\big(X^{(1:n)}_t\big)$ and compute
\begin{equation}
g\big(X^{(i)}_t, X^{(-i)}_t, t\big) = \nabla_{X^{(i)}_t}\, h\big(X^{(1:n)}_t\big)
\end{equation}
for each $i$. Adding a normalized version of $g$ as a diversity velocity increases $h$ during integration. The core trade-off remains: strong diversity improves spread but risks low-density drift; weak diversity preserves quality but limits gains.

\paragraph{Keeping samples within the data manifold.}
A diversity-driven move $x_t \mapsto x_t+\delta x$ departs from the manifold if it reduces the log-density at time $t$, \ie, $\log p_t(x_t+\delta x) < \log p_t(x_t)$. To avoid this, constrain the move to satisfy
\begin{equation}
    \delta x \cdot s(x,t) \ge 0,
\end{equation}
where $s(x,t):=\nabla_x \log p_t(x)$ is the score. Decompose
\begin{equation}
    g \;=\; g_{\parallel} + g_{\perp},
\end{equation}
with respect to the unit score direction $\hat{s}=s/\|s\|_2$, where $g_{\parallel}=(g\cdot \hat{s})\,\hat{s}$ and $g_{\perp}=g-g_{\parallel}$. 
Regularize as
\begin{equation}
    g_\mathrm{reg} = \alpha(t)\, g_{\parallel} + g_{\perp},
\end{equation}
with $\alpha(t)=1$ when $g\cdot \hat{s}\ge 0$ (toward higher density). When $g\cdot \hat{s}<0$ (toward off-manifold), we consider:
\begin{itemize}
    \item \emph{Soft}: $\alpha(t)=1-t$ (weak early, strong near $t=1$);
    \item \emph{Hard}: $\alpha(t)=0$ (remove the harmful component).
\end{itemize}
For rectified flows, the score is computed directly from $v$ via Eq.~(\ref{eq:score_velocity}). Since $v(x,t)$ is already computed at each step, applying score-based regularization incurs negligible additional overhead. After computing $g_\mathrm{reg}$ for all samples and jointly normalizing them, we obtain diversity velocities $u$ with \emph{amplified on-manifold components} and reduced (or removed) off-manifold components.

\paragraph{Illustration.}
In Figure~\ref{fig:diversity_force}, we show two joint samples $x_t^1$ and $x_t^2$. Since the scores satisfy $s_t^1\cdot g_t^1>0$ and $s_t^2\cdot g_t^2<0$, regularization is active only for the second sample. Thus, $(g_t^1)_\mathrm{reg}=g_t^1$ and $(g_t^2)_\mathrm{reg}=(g_t^2)_\perp$ (the hard version). After normalization, the diversity velocities $u_1,u_2$ push samples apart while avoiding departures from the manifold.

\subsection{Importance Weight Estimation}
\label{subsec:method_importance}

\paragraph{Representing the non-IID marginal with a residual velocity.}
To obtain unbiased estimates, we compute importance weights for non-IID, jointly sampled outputs. Because the joint draw $X^{(1:n)}$ is realized once, the induced single-sample marginal $p'(x)$ is difficult to evaluate directly. We learn a lightweight \emph{residual velocity} $r_\phi(x,t)$ such that the perturbed flow
\begin{equation}
    \dot{X}_t = v(X_t,t) + r_\phi(X_t,t), \quad X_0 \sim p_0,
\end{equation}
matches the \emph{marginal} distribution induced by the diversity-coupled sampler at $t=1$. Since rectified-flow models~\citep{DBLP:conf/iclr/LiuG023} are widely used~\citep{DBLP:conf/icml/EsserKBEMSLLSBP24,labs2025flux1kontextflowmatching,flux2024}, we focus on rectified flows, where learning such a residual velocity field is sufficient for importance-weight estimation, and a small network (compared to that for $v$) can be used to reduce training and inference cost.

Our method is also suitable for general flow matching models, but score functions then need to be learned, \eg, via score matching~\citep{DBLP:conf/iclr/0011SKKEP21}. 
We summarize the three ODEs used throughout: the standard Original ODE (Definition~\ref{def:original_ode}), the Joint ODE we use for non-IID sampling (Definition~\ref{def:joint_ode}), and our learned Marginal ODE for importance weighting (Definition~\ref{def:marginal_ode}).

\begin{defbox}
\begin{definition}[Original ODE]\label{def:original_ode}
\[
  \dot X_t = v(X_t,t), \quad X_0\sim p_0.
\]
We denote its probability density as \(p_t(x)\).
\end{definition}

\begin{definition}[Joint ODEs]\label{def:joint_ode}
\[
  \dot X_t^{(i)} = v \big(X_t^{(i)}, t\big)
  + u\big(X^{(i)}_t,X^{(-i)}_t,t\big), \quad
  X_0^{(i)} \overset{\mathrm{IID}}{\sim} p_0.
\]
We denote its marginal probability density as \(p'_t(x)\).
We use this setup for sampling.
\end{definition}

\begin{definition}[Marginal ODE]\label{def:marginal_ode}
\[
  \dot X_t = v(X_t,t) + r_\phi(X_t,t), \quad X_0\sim p_0.
\]
We denote its probability density as \(p''_{\phi,t}(x)\) and its score as \(s''_\phi(x,t):=\nabla_x\log p''_{\phi,t}(x)\).
We train \(r_\phi\) such that
\(p''_{\phi,1}(x) \approx p'_1(x)\) for \(x\)
generated by the joint ODEs.
\end{definition}
\end{defbox}

\paragraph{Density evolution along the sampling path.}
We aim to compute the importance weight
\begin{equation}
\begin{aligned}
    w(x) &
    = \frac{p_1(x)}{p'_1(x)}  \approx \frac{p_1(x)}{p''_{\phi,1}(x)}.
\end{aligned}
\end{equation}
To compute this, we define a time-dependent ratio
\begin{equation}
    w_{\phi,t}(x) \;:=\; \frac{p_t(x)}{p''_{\phi,t}(x)},
\end{equation}
so that $w_{\phi,1}(x) \approx w(x)$ and $w_{\phi,0}(x) = 1$.
From Eq.~(\ref{eq:continuity}), we derive the evolution of $\log w_{\phi,t}(x)$ at a fixed position $x$, as given in Theorem~\ref{theorem:log_w_fixed}.
Since sampling proceeds along the coupled dynamics in Definition~\ref{def:joint_ode},
\begin{equation}
\dot X_t^{(i)} = v(X_t^{(i)}, t) + u(X_t^{(i)}, X_t^{(-i)}, t), \!\!\quad\!\! i = 1, \dots, n,
\end{equation}
combining this with Theorem~\ref{theorem:log_w_fixed} yields Theorem~\ref{theorem:log_w_along},
which characterizes how $w_{\phi,t}(X_t^{(i)})$ evolves along each trajectory.
As discussed in Section~\ref{subsec:design_discussion}, estimating the density ratio along trajectories addresses key limitations of fixed-position estimation, and we therefore adopt this as our default method.
For rectified flows, the computation simplifies further---requiring no additional score function estimation---as shown in Corollary~\ref{corollary:log_w_along_rf}.
All proofs are provided in Appendix~\ref{app:evolution_proofs}.

\begin{thmbox}
\begin{theorem}[Evolution of the Importance Weight at Fixed Position]
\label{theorem:log_w_fixed}
At a fixed position $x$, the evolution of $\log w_{\phi,t}(x) \!=\! \log p_t(x) \!-\! \log p''_{\phi,t}(x)$ is 
\[
\begin{aligned}
    \partial_t \log w_{\phi,t}(x)
    & = \nabla_x \cdot r_\phi(x,t) + s''_\phi(x,t) \cdot r_\phi(x,t) \\
    & \quad + v(x,t) \cdot \big(s''_\phi(x,t) - s(x,t)\big).
\end{aligned}
\]
\end{theorem}

Intuitively, this tracks the density ratio at a fixed point $x$ as time advances; from $w_{\phi,0}(x)=1$, integrating to $t=1$ recovers the weight $w(x)$.

\begin{theorem}[Evolution of the Importance Weight Along the Trajectory]
\label{theorem:log_w_along}
Along the sample trajectory, the evolution of $\log w_{\phi,t}(X_t^{(i)})$ is
\[
\begin{aligned}
& \frac{\mathrm{d}}{\mathrm{dt}} \log w_{\phi,t}(X_t^{(i)}) \\
& = \nabla_x \cdot r_\phi(X_t^{(i)},t) + s''_\phi(X_t^{(i)},t) \cdot r_\phi(X_t^{(i)},t) \\
& - \big(s_\phi''(X_t^{(i)},t) - s(X_t^{(i)},t)\big) \cdot u(X^{(i)}_t,X^{(-i)}_t,t).
\end{aligned}
\]
\end{theorem}

Here the ratio is tracked along each sample's own trajectory. We adopt this as our default estimator, since integrating along the sampled path avoids evaluating $r_\phi$ off its training distribution (Section~\ref{subsec:design_discussion}).
\end{thmbox}

\begin{corbox}
\begin{corollary}[Evolution under Rectified Flows]
\label{corollary:log_w_along_rf}
Suppose $v$ and $v+r_\phi$ are rectified flows, then Theorem~\ref{theorem:log_w_along} simplifies without learning additional score functions:
\[
\begin{aligned}
& \frac{\mathrm{d}}{\mathrm{dt}} \log w_{\phi,t}(X_t^{(i)}) \\
& = \nabla_x \cdot r_\phi(X_t^{(i)},t) + \tfrac{t v(X^{(i)}_t,t) - X^{(i)}_t}{1-t} \cdot r_\phi(X_t^{(i)},t) \\
& + \tfrac{t}{1-t} \big(r_\phi(X_t^{(i)},t) - u\big(X^{(i)}_t,X^{(-i)}_t,t\big)\big) \cdot r_\phi(X_t^{(i)},t).
\end{aligned}
\]
\end{corollary}
\end{corbox}

\paragraph{Learning the residual velocity.}
We train a \emph{rectified flow} to represent the non-IID marginal induced by the diversity-enhanced ODEs. Rather than re-learning $v$, we learn only the residual $r_\phi$, which together with $v$ forms the marginal velocity $v+r_\phi$. We adopt the rectified-flow objective
\begin{equation}
    \min_\phi \mathbb{E}_{X_0, X_1,t}\Big[\big\|(X_1 - X_0) - v(X_t, t) - r_\phi(X_t, t)\big\|^2_2\Big],
\end{equation}
where $X_0 \sim p_0$, $X_1$ is sampled from the diversity-enhanced non-IID sampler, and $X_t = (1-t)\,X_0 + t\,X_1$. 
The training target does not require external labels since it is created by our sampler.
In practice, a relatively lightweight neural network for $r_\phi$ can be used to accelerate both training and inference, because it is based on $v$ (See Appendix~\ref{app:rphi_arch} for more details).

\subsection{Discussions}

\label{subsec:design_discussion}

We discuss two variations to justify our design choices.

\paragraph{Variation 1. Directly sampling from the marginal velocity.}
Since $r_\phi$ is trained, one can sample independently from $v+r_\phi$ (the ODE in Definition~\ref{def:marginal_ode}), which matches the marginal distribution of diversity-enhanced samples. However, IID batches drawn from $v+r_\phi$ are typically less diverse than those generated by the coupled ODEs in Definition~\ref{def:joint_ode}, which explicitly enhance diversity across concurrent trajectories (see Table~\ref{tab:exp_gaussian_diversity_quality}).

\paragraph{Variation 2. Estimating importance weights via fixed-position evolution (for samples at $t=1$).}
In Theorem~\ref{theorem:log_w_fixed}, we estimate the importance weight at the fixed position for each sample as
\begin{equation}
\begin{aligned}
    \log w_{\phi,1}(x)
    & \!=\! \int_0^1 \!\Big( \nabla_x \cdot r_\phi(x,t) \!+ \!s''_\phi(x,t) \cdot r_\phi(x,t)\\
    & \quad + v(x,t) \cdot \big(s''_\phi(x,t) - s(x,t)\big) \Big) \mathrm{dt},
\end{aligned}
\end{equation}
which, given $v$ and $v+r_\phi$ are rectified flows, further simplifies (Proof in Appendix~\ref{app:log_w_fixed_rf}):
\begin{equation}
\begin{aligned}
    \log w_{\phi,1}(x)
    & \!=\! \int_0^1 \!\Big(\nabla_x \cdot r_\phi(x,t) + \frac{1}{1-t} r_\phi(x,t) \cdot \\
    & \big(2 t v(x,t) + t r_\phi(x,t) - x\big) \Big) \mathrm{dt}.
\label{eq:log_w_evolution_fixed}
\end{aligned}
\end{equation}
However, in practice, the pair $(X_1^{(i)},t)$ might be unlikely to occur along the real sampling path when $t<1$, because $X_1^{(i)}$ is the sample for time $t=1$. In conditional generation, \ie, generating an image for the text ``a cat'', $\mathbb{E}X_1^{(i)}\neq\mathbf{0}$ but $\mathbb{E}X_0^{(i)}=\mathbf{0}$, which means $X_1^{(i)}$ can be rare among real samples at $t=0$, \ie, $p_0(X_1^{(i)})\approx 0$. This implies that the training data of $r_\phi$ cannot cover this situation, potentially leading to an undefined output $r_\phi(X_1^{(i)},t)$ for small $t$ and worsening performance. Our trajectory-based estimator instead integrates \emph{along the actual path} and avoids out-of-distribution inputs to $r_\phi$ (see Table~\ref{tab:exp_gaussian_mixture_density_comparison}).

\section{Experiments}
\label{sec:experiments}

We first evaluate on a \textbf{Gaussian mixture} with known density, enabling precise diagnosis of diversity, quality, and estimation of importance weights and expectations. We then assess \emph{plug-in} benefits on large flow models---\textbf{Stable Diffusion 3.5 Medium}~\citep{DBLP:conf/icml/EsserKBEMSLLSBP24} for text-to-image generation and \textbf{FLUX.1-Fill-dev}~\citep{labs2025flux1kontextflowmatching,flux2024} for image inpainting---to study distribution coverage in complex image settings. Diversity velocities and implementation details for each experiment are in Appendix~\ref{app:diversity_velocities} and Appendix~\ref{app:exp_details}, respectively.

\subsection{Experiments on Gaussian Mixture}

\label{subsec:exp_gaussian_mixture}

The experiment uses a mixture of ten Gaussians with centers uniformly placed on a unit circle in the first two dimensions. Each trial jointly samples ten points.

\subsubsection{Sample Diversity and Quality}

\paragraph{Setup.}
We use a sparse $8$-D near-planar distribution with variances of $0.01$ on the first two coordinates and $0.0001$ on the remaining six (mimicking a sparse data distribution).
We run 10{,}000 trials, evaluating several diversity objectives and testing whether adding our \emph{score-based regularization} (SR) improves performance.
For \textit{diversity}, we assign each sample to its nearest Gaussian and report \textit{mode coverage}: the number of distinct modes captured by the 10 joint samples. We also report a \textit{Marginal} variant by pooling joint samples across repetitions, randomly regrouping them into sets of 10, and computing the resulting coverage---this mimics IID sampling from the \emph{marginal} of the diversity-enhanced sampler.
For \textit{quality}, we report $\log p$ (density at each sample computed under the original Gaussian mixture) and RMSE to the nearest mode.

\paragraph{Results.}
In Table~\ref{tab:exp_gaussian_diversity_quality}, across all diversity objectives, adding SR (soft or hard) \emph{significantly} improves quality ($\log p$ increases; RMSE decreases) while preserving joint mode coverage. For example, with DPP, $\log p$ and RMSE improve from $7.16$ and $0.527$ to $19.27$ and $0.253$ under SR-hard, while joint coverage remains almost unchanged ($8.55{\to}8.57$). By contrast, DiverseFlow (DPP$+$DF) improves quality but \emph{reduces} diversity ($8.55{\to}6.89$), evidencing a diversity--quality trade-off that our score-based regularization alleviates.
Finally, IID sampling from the diversity-enhanced \emph{marginal} fails to recover joint-diversity benefits (\eg, DPP joint coverage $8.55$ vs.\ marginal $6.51$, close to original IID $6.51$), justifying our design choice of \emph{joint} sampling via coupled ODEs (Definition~\ref{def:joint_ode}) rather than IID from the diversity-enhanced marginal (Definition~\ref{def:marginal_ode}).

\begin{table}[t]
\centering
\setlength{\tabcolsep}{3pt}
\begin{tabular}{l|c|cccc}
\toprule
\multirow{2}{*}{Method} & \multirow{2}{*}{SR} & \multicolumn{2}{c}{Mode coverage$\uparrow$} & \multirow{2}{*}{$\log p$ $\uparrow$} & \multirow{2}{*}{RMSE $\downarrow$} \\
&&Joint & Marginal&& \\
\midrule
IID & - & - & \color{ignore}{6.51(2)} & 20.31(1) & 0.139(1)\\
\midrule
\multirow{3}{*}{PG} & \no & 7.50(2) & \color{ignore}{6.51(2)} & 7.28(7) & 0.526(1) \\
& soft & 7.56(2) & \color{ignore}{6.51(2)} & 14.94(4) & 0.386(1) \\
& hard & \textbf{7.59(2)} & \color{ignore}{6.52(2)} & \textbf{19.93(1)} & \textbf{0.227(1)} \\
\midrule
\multirow{3}{*}{Chebyshev} & \no & 8.39(1) & \color{ignore}{6.51(2)} & 7.74(6) & 0.517(1) \\
& soft & \textbf{8.60(1)} & \color{ignore}{6.52(2)} & 15.15(3) & 0.379(1) \\
& hard & 8.58(1) & \color{ignore}{6.52(2)} & \textbf{19.90(1)} & \textbf{0.227(1)} \\
\midrule
\multirow{3}{*}{Reciprocal} & \no & 9.45(1) & \color{ignore}{6.51(2)} & 7.91(6) & 0.514(1) \\
& soft & \textbf{9.55(1)} & \color{ignore}{6.51(2)} & 14.71(4) & 0.390(1) \\
& hard & \textbf{9.55(1)} & \color{ignore}{6.51(2)} & \textbf{19.34(2)} & \textbf{0.247(1)} \\
\midrule
\multirow{3}{*}{Log-barrier} & \no & 9.44(1) & \color{ignore}{6.52(2)} & 7.50(7) & 0.521(1) \\
& soft & 9.62(1) & \color{ignore}{6.53(2)} & 15.08(3) & 0.380(1) \\
& hard & \textbf{9.63(1)} & \color{ignore}{6.52(2)} & \textbf{19.84(1)} & \textbf{0.227(1)} \\
\midrule
DPP$+$DF & - & 6.89(2) & \color{ignore}{6.50(2)} & \textbf{20.96(1)} & \textbf{0.121(1)} \\
\multirow{3}{*}{DPP} & \no & 8.55(2) & \color{ignore}{6.51(2)} & 7.16(6) & 0.527(1) \\
& soft & 8.56(2) & \color{ignore}{6.50(2)} & 14.25(3) & 0.402(1) \\
& hard & \textbf{8.57(2)} & \color{ignore}{6.50(2)} & 19.27(1) & 0.253(1) \\
\bottomrule
\end{tabular}
\caption{\textbf{Sampling diversity and quality results for the Gaussian mixture experiment.} Mode coverage, $\log p$, and RMSE are reported as mean(uncertainty) with 95\% confidence intervals. SR (soft and hard) is our score-based regularization, applied on top of each diversity method. For each metric, the best result within each method group is shown in \textbf{bold}; the IID reference is excluded.}
\label{tab:exp_gaussian_diversity_quality}
\end{table}

\begin{table}[t]
\centering
\begin{tabular}{l|cccc}
\toprule
 & SE $\downarrow$ & $\tau_b$ $\uparrow$ & $\rho$ $\uparrow$ & gAP $\uparrow$ \\
\midrule
KDE & 8.0(1) & 0.348(6) & 0.449(7) & 0.736(3) \\
kNN & 9.9(1) & 0.327(6) & 0.424(8) & 0.734(3) \\
MGF & 8.8(1) & 0.327(6) & 0.427(8) & 0.736(3) \\
\midrule
Ours$^{\dagger}$ & 3.8(1) & 0.498(5) & 0.621(6) & 0.794(2) \\
Ours & \textbf{2.6(1)} & \textbf{0.548(5)} & \textbf{0.668(6)} & \textbf{0.805(2)} \\
\bottomrule
\end{tabular}
\caption{\textbf{Importance-weight estimation results for the Gaussian mixture experiment.} SE and ranking metrics are reported as mean(uncertainty) with 95\% confidence intervals. $^{\dagger}$ indicates the variation of our method that estimates the importance-weight evolution at a fixed, final sample position. The best-performing method for each metric is shown in \textbf{bold}.}
\label{tab:exp_gaussian_mixture_density_comparison}
\end{table}

\subsubsection{Importance Weight \& Expectation Estimation}

\paragraph{Setup.}
For accurate ground truth, we use a 2D Gaussian mixture, shifted by $1$ along the first axis to induce a moderate distribution shift, where each component $k$ has a weight proportional to $2^{-k}$. Non-IID joint sets are sampled using a Harmonic DPP with soft SR. The ground-truth importance weights are computed as the ratio between the original density and the non-IID marginal estimated via Local-Likelihood Density Estimation (LLDE)~\citep{Loader1996,HjortJones1996} (see Appendix~\ref{app:llde_details}). We train the residual velocity $r_\phi$ on $1{,}000$ joint sets (each containing $10$ trajectories) and evaluate on $10{,}000$ sets. We compare three types of estimators: (i) \emph{trajectory-based} evolution (ours, Theorem~\ref{theorem:log_w_along}); (ii) \emph{fixed-position} evolution at a terminal state (Theorem~\ref{theorem:log_w_fixed}); and (iii) density baselines---k-Nearest Neighbors Density Estimation (kNN)~\citep{loftsgaarden1965nonparametric}, Gaussian Kernel Density Estimation (KDE)~\citep{DBLP:books/sp/Silverman86}, and Multivariate Gaussian Fit (MGF)~\citep{DBLP:journals/jei/BishopN07} (see Appendix~\ref{app:baselines}).

\paragraph{Importance-weight accuracy.}
We report squared error (SE) and ranking metrics---Kendall's $\tau_b$~\citep{kendall1945treatment}, Spearman's $\rho$~\citep{spearman1987proof}, and graded AP (gAP)~\citep{DBLP:conf/sigir/MoffatM0A22} (see Appendix~\ref{app:ranking_metrics})---computed on the top $50\%$ most-confident ground truth. In Table~\ref{tab:exp_gaussian_mixture_density_comparison}, density baselines exhibit large SE and weak ranking, while the \emph{trajectory-based} estimator consistently outperforms the \emph{fixed-position} variant, since path integration avoids out-of-distribution inputs to $r_\phi$.

\paragraph{Function expectation.}
We estimate $\mu{=}\mathbb{E} f(X)$, where $f(x)$ maps each sample to a one-hot vector indicating the index of its nearest Gaussian. Ground-truth $\mu$ is computed from $1{,}000{,}000$ IID samples drawn from the original flow-matching model, yielding the probability that a generated sample belongs to each Gaussian. We evaluate on joint samples via Jensen--Shannon (JS) divergence~\citep{DBLP:journals/tit/Lin91}, which remains well-defined when some predicted probabilities are zero. Our \emph{trajectory-based} weighting debiases the equal-weighted non-IID average (JS $0.088{\to}0.071$) and improves over spending the same budget on additional IID draws ($0.077$); the \emph{fixed-position} variant ($0.078$) lies between. The density baselines attain the lowest JS here, but this is potentially because they are additionally given the analytic, closed-form target density~$p_1$, the true density our estimator never accesses; the full comparison is in Appendix~\ref{app:expectation_full}.

\paragraph{Summary.}
Our along-trajectory estimator outperforms its fixed-position variant and the density baselines in importance-weight estimation; used downstream, these weights debias the equal-weighted non-IID average and improve over additional IID sampling.

\begin{table*}[t]
\centering
\begin{tabular}{l|c|cccccc|c}
\toprule
 & SR & $T_1$ & $T_2$ & $T_3$ & $T_4$ & $T_5$ & $T_6$ & Mean \\
\midrule
IID & - & 22.6(4) & 22.5(4) & 18.6(2) & 19.1(2) & 20.5(2) & 22.0(2) & 20.9(2) \\
\midrule
\multirow{3}{*}{DPP} & \no & 19.2(3) & 18.8(4) & 15.3(2) & 15.7(2) & 17.5(2) & 18.8(2) & 17.5(2) \\
 & soft & 19.0(3) & 18.6(4) & 15.2(2) & 15.6(2) & 17.4(2) & 18.7(2) & 17.4(2) \\
 & hard & \textbf{18.7(3)} & \textbf{18.4(4)} & \textbf{15.1(2)} & \textbf{15.5(2)} & \textbf{17.2(2)} & \textbf{18.4(2)} & \textbf{17.2(2)} \\
\midrule
\multirow{3}{*}{PG} & \no & 18.1(3) & 17.0(3) & 14.5(2) & 14.9(2) & 17.6(2) & 19.1(2) & 16.9(2) \\
 & soft & 18.0(3) & 16.9(3) & 14.5(2) & 14.8(2) & 17.6(2) & 19.0(2) & 16.8(2) \\
 & hard & \textbf{17.7(3)} & \textbf{16.7(3)} & \textbf{14.4(2)} & \textbf{14.7(2)} & \textbf{17.4(2)} & \textbf{18.8(2)} & \textbf{16.6(2)} \\
\midrule
\multirow{3}{*}{DF} & \no & 18.9(3) & 18.6(4) & 15.2(2) & 15.6(2) & 17.4(2) & 18.6(2) & 17.4(2) \\
 & soft & 18.9(3) & 18.5(4) & 15.1(2) & 15.5(2) & 17.3(2) & 18.6(2) & 17.3(2) \\
 & hard & \textbf{18.7(3)} & \textbf{18.4(4)} & \textbf{15.1(2)} & \textbf{15.5(2)} & \textbf{17.2(2)} & \textbf{18.4(2)} & \textbf{17.2(2)} \\
\bottomrule
\end{tabular}
\caption{\textbf{Coverage radius results on text-to-image generation.} Values are mean(uncertainty) with 95\% confidence intervals. SR (soft and hard) is our score-based regularization, applied on top of each diversity method.}
\label{tab:text2image}
\end{table*}

\begin{table*}[t]
\centering
\setlength{\tabcolsep}{4.5pt}
\begin{tabular}{l|c|cccccccccc|c}
\toprule
Method & SR & $T_1$ & $T_2$ & $T_3$ & $T_4$ & $T_5$ & $T_6$ & $T_7$ & $T_8$ & $T_9$ & $T_{10}$ & Mean \\
\midrule
IID & - & 2.47(4) & 0.22(1) & 0.09(1) & 0.15(1) & 0.46(2) & 1.43(5) & 0.07(1) & 0.09(1) & 0.25(1) & 0.11(1) & 0.53(1) \\
\midrule
\multirow{3}{*}{DPP} & \no & 2.20(4) & 0.41(1) & 0.09(1) & 0.17(1) & 0.35(1) & 0.95(5) & 0.21(1) & 0.13(1) & 0.28(1) & 0.17(1) & 0.50(1) \\
& soft & \textbf{2.19(4)} & 0.38(1) & 0.09(1) & 0.16(1) & 0.33(1) & \textbf{0.89(4)} & 0.20(1) & 0.12(1) & 0.25(1) & 0.16(1) & 0.48(1) \\
& hard & 2.26(3) & \textbf{0.17(1)} & \textbf{0.07(1)} & \textbf{0.11(1)} & \textbf{0.30(2)} & 1.11(5) & \textbf{0.10(1)} & \textbf{0.08(1)} & \textbf{0.19(1)} & \textbf{0.09(1)} & \textbf{0.45(1)} \\
\midrule
\multirow{3}{*}{PG} & \no & 2.14(3) & 0.53(2) & 0.11(1) & 0.21(1) & 0.49(2) & 0.63(4) & 0.25(1) & 0.16(1) & 0.34(1) & 0.20(1) & 0.51(1) \\
& soft & \textbf{2.13(3)} & 0.52(2) & 0.10(1) & 0.21(1) & 0.47(2) & \textbf{0.63(4)} & 0.24(1) & 0.15(1) & 0.32(1) & 0.19(1) & 0.50(1) \\
& hard & 2.21(3) & \textbf{0.21(1)} & \textbf{0.08(1)} & \textbf{0.15(1)} & \textbf{0.32(1)} & 0.93(5) & \textbf{0.15(1)} & \textbf{0.10(1)} & \textbf{0.18(1)} & \textbf{0.10(1)} & \textbf{0.44(1)} \\
\midrule
\multirow{3}{*}{DF} & \no & 2.20(4) & 0.40(1) & 0.09(1) & 0.17(1) & 0.34(1) & 0.92(4) & 0.21(1) & 0.13(1) & 0.27(1) & 0.17(1) & 0.49(1) \\
& soft & \textbf{2.19(4)} & 0.36(1) & 0.09(1) & 0.16(1) & 0.32(1) & \textbf{0.89(5)} & 0.19(1) & 0.12(1) & 0.24(1) & 0.16(1) & 0.47(1) \\
& hard & 2.24(3) & \textbf{0.17(1)} & \textbf{0.07(1)} & \textbf{0.11(1)} & \textbf{0.30(2)} & 1.10(5) & \textbf{0.10(1)} & \textbf{0.08(1)} & \textbf{0.20(1)} & \textbf{0.09(1)} & \textbf{0.45(1)} \\
\bottomrule
\end{tabular}
\caption{\textbf{Coverage radius results on image inpainting.} Values are mean(uncertainty) with 95\% confidence intervals. SR (soft and hard) is our score-based regularization, applied on top of each diversity method.}
\label{tab:inpainting}
\end{table*}

\subsection{Experiments on Text-to-Image Generation}

\label{subsec:exp_text_to_image}

We next test Stable Diffusion 3.5 Medium~\citep{DBLP:conf/icml/EsserKBEMSLLSBP24} for text-to-image generation.
We study whether using diverse samples can better cover the IID samples $S_\text{IID}$ with a fixed budget of samples $S_\text{Rep}$, measured by the \textbf{coverage radius} 
\begin{equation}
\mathcal{R}(S_\text{IID}|S_\text{Rep}) = \max_{I_1 \in S_\text{IID}}
\min_{I_2 \in S_\text{Rep}} d^2(I_1, I_2), 
\label{eq:representation_error}
\end{equation}
where $d(I_1, I_2)$ is the distance between two images; to avoid evaluation bias, we do not introduce any embedding model and instead use the distance between their spatial means in latent space. This metric quantifies worst-case geometric coverage in the latent-space representation, which is more meaningful than a perceptual score.
The text prompts are:
\noindent
\begin{minipage}[t]{0.5\linewidth}
\begin{itemize}
    \item $T_1$: ``a fish''
    \item $T_3$: ``a cat''
    \item $T_5$: ``something outside''
\end{itemize}
\end{minipage}\hfill
\begin{minipage}[t]{0.5\linewidth}
\begin{itemize}
    \item $T_2$: ``a realistic fish''
    \item $T_4$: ``a releastic cat''
    \item $T_6$: ``someone ahead''
\end{itemize}
\end{minipage}
They cover different scenarios: $T_1{\to}T_2$ (coarse$\to$fine conditions), $T_3{\to}T_4$ (low-quality conditions, \eg, typos), and $T_5$, $T_6$ (ambiguous conditions). For each prompt, we generate $10{,}000$ images for $S_\text{IID}$ and $10$ images for each $S_\text{Rep}$ ($1{,}000$ repetitions) under evaluation.

\begin{figure}[t]
\centering
\includegraphics[width=\linewidth]{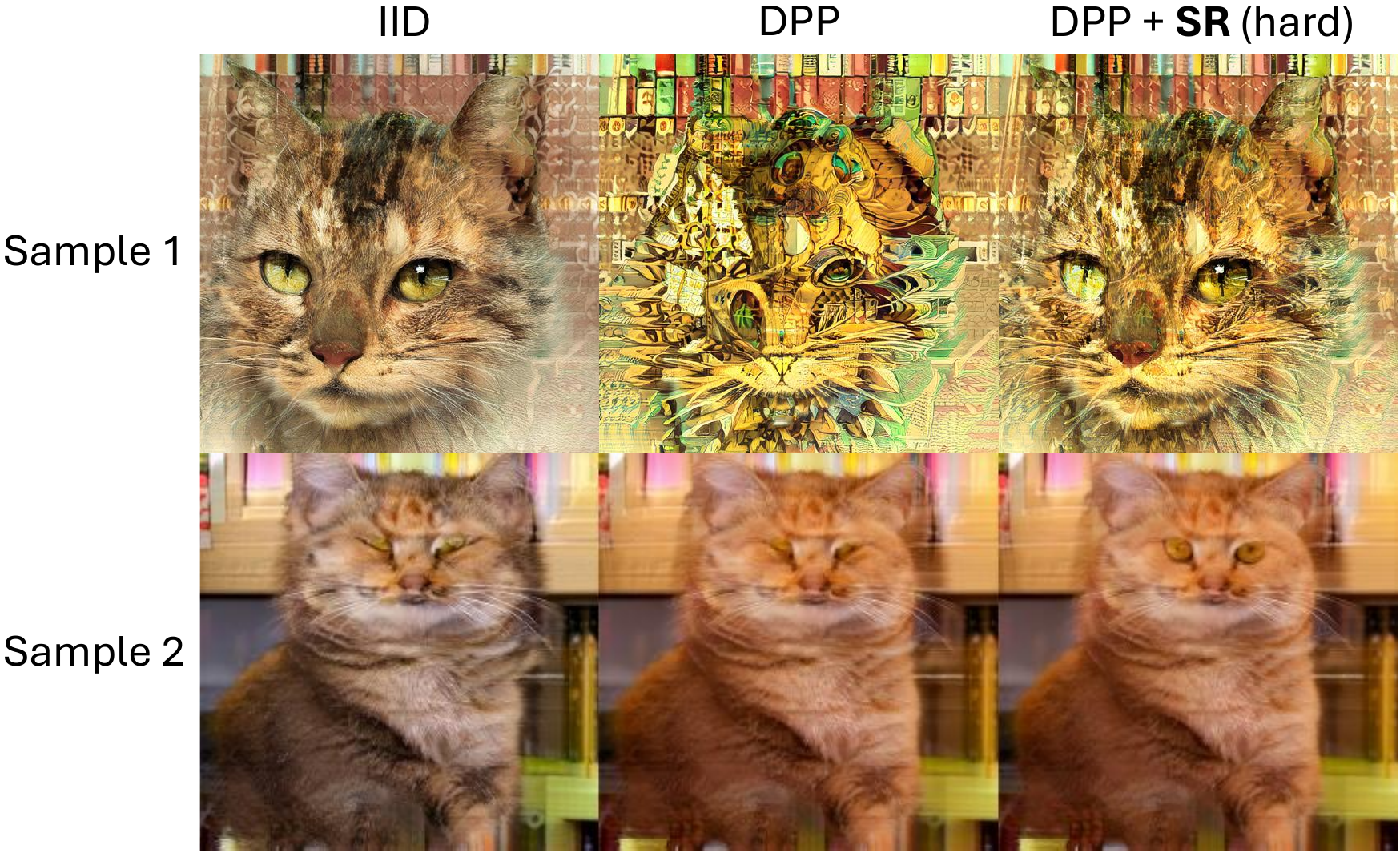}
\caption{
\textbf{Qualitative results for text-to-image generation.}
Although diverse, \textbf{Sample 1} from DPP appears unreasonable. Adding SR (hard) preserves diversity while making it reasonable. For \textbf{Sample 2}, SR (hard) refines the cat's eyes.
}
\label{fig:text2image}
\end{figure}

Table~\ref{tab:text2image} shows that introducing diversity terms reduces the coverage radius for all prompts $T_1$ to $T_6$ relative to IID (taking $S_\text{Rep}$ as 10 IID samples), indicating improved coverage under a fixed budget. Adding our score-based regularization (SR) further improves performance across all prompts, suggesting that maintaining on-manifold quality increases sample efficiency (fewer ``wasted'' samples). Qualitative results are shown in Figure~\ref{fig:text2image}.

\subsection{Experiments on Image Inpainting}

\label{subsec:exp_inpainting}

We additionally evaluate performance on image inpainting, which is more constrained than text-to-image, to test whether improvements persist. We use FLUX.1-Fill-dev~\citep{labs2025flux1kontextflowmatching,flux2024}. We select 10 square, high-quality images from the test split of the Open Images Dataset V7~\citep{DBLP:journals/ijcv/KuznetsovaRAUKP20} and apply a mask with four squares to inpaint. For each image, we generate $|S_\text{IID}|\!=\!1{,}000$ inpaintings and evaluate $|S_\text{Rep}|\!=\!10$ jointly sampled inpaintings (100 repetitions). The coverage radius (Eq.~(\ref{eq:representation_error})) is computed over average latent features \emph{within the masked regions}.

\begin{figure}[t]
    \centering
    \includegraphics[width=\linewidth]{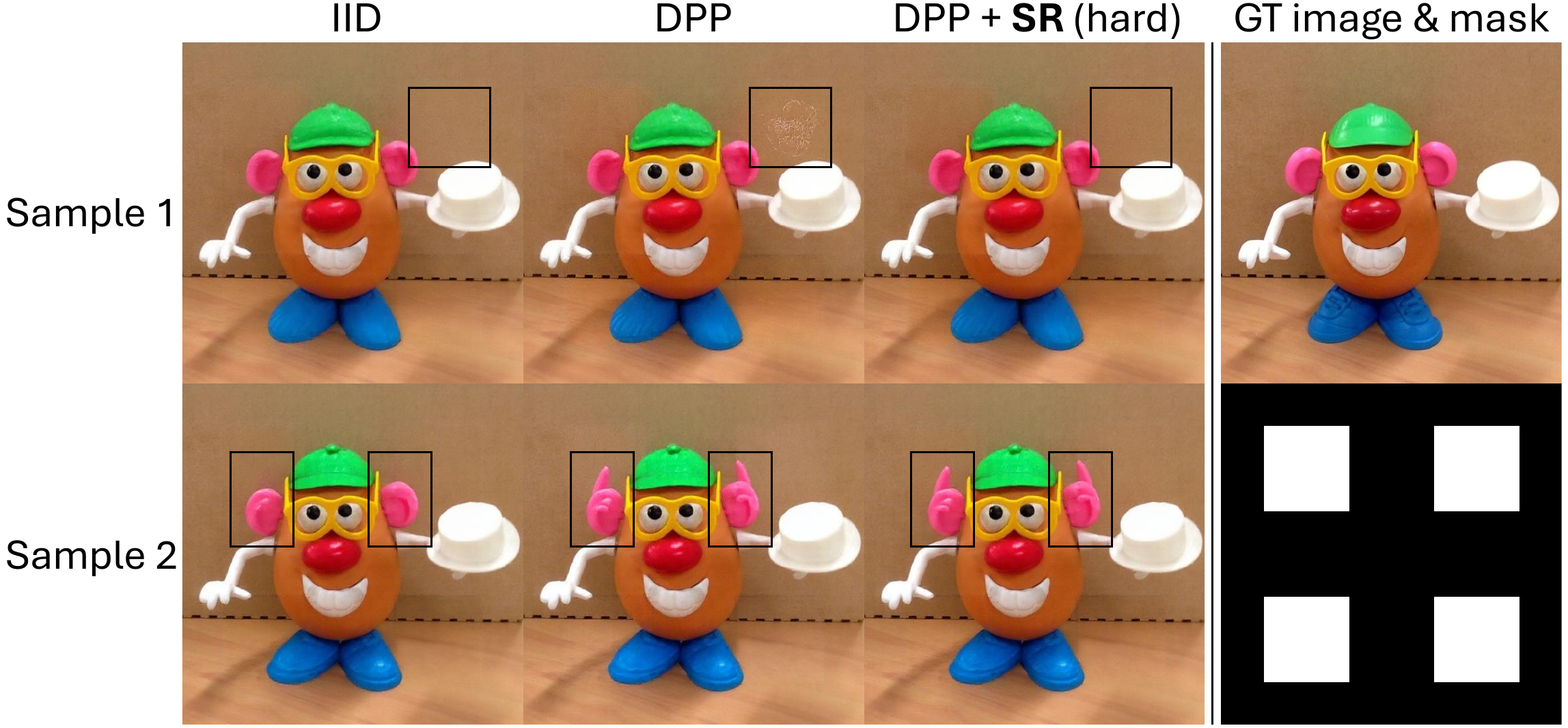}
\caption{
\textbf{Qualitative results for image inpainting.} 
Two samples are shown from the joint samples, with all methods sharing identical initialization. In \textbf{Sample 1}, DPP introduces artifacts (highlighted by the black rectangle) that are removed by SR (hard). While enhancing quality, DPP$+$SR retains the diversity of DPP, as illustrated in \textbf{Sample 2}.}
\label{fig:inpainting}
\end{figure}

Table~\ref{tab:inpainting} shows that diversity terms improve the mean coverage radius across images relative to IID. Similar to text-to-image generation, adding SR improves the mean coverage radius for every method, though SR-hard can regress on a few inputs (\eg, $T_1$ and $T_6$), indicating that maintaining on-manifold quality increases sample efficiency.
Qualitative results are shown in Figure~\ref{fig:inpainting}.

\section{Conclusion}

We introduced an \emph{importance-weighted non-IID sampling} framework for flow matching that (i) applies \emph{score-based regularization for diversity velocity} to push concurrent trajectories apart while keeping them on-manifold, and (ii) computes \emph{importance weights} along trajectories by learning a residual velocity. On a Gaussian mixture, SR improves sample quality while preserving diversity, and our importance weighting yields the most accurate importance-weight estimates, improving expectation estimation over both equal weighting and additional IID sampling. On Stable Diffusion 3.5 Medium and FLUX.1-Fill-dev, diversity with SR consistently reduces the mean sample coverage radius.

\bibliography{ref}

\begin{thebibliography}{31}
\providecommand{\natexlab}[1]{#1}
\providecommand{\url}[1]{\texttt{#1}}
\expandafter\ifx\csname urlstyle\endcsname\relax
  \providecommand{\doi}[1]{doi: #1}\else
  \providecommand{\doi}{doi: \begingroup \urlstyle{rm}\Url}\fi

\bibitem[Bishop and Nasrabadi(2007)]{DBLP:journals/jei/BishopN07}
Christopher~M. Bishop and Nasser~M. Nasrabadi.
\newblock \emph{Pattern Recognition and Machine Learning}.
\newblock \emph{J. Electronic Imaging}, 16\penalty0 (4):\penalty0 049901, 2007.
\newblock \doi{10.1117/1.2819119}.
\newblock URL \url{https://doi.org/10.1117/1.2819119}.

\bibitem[Corso et~al.(2024)Corso, Xu, Bortoli, Barzilay, and Jaakkola]{DBLP:conf/iclr/CorsoXBBJ24}
Gabriele Corso, Yilun Xu, Valentin~De Bortoli, Regina Barzilay, and Tommi~S. Jaakkola.
\newblock Particle guidance: non-i.i.d. diverse sampling with diffusion models.
\newblock In \emph{The Twelfth International Conference on Learning Representations, {ICLR} 2024, Vienna, Austria, May 7-11, 2024}. OpenReview.net, 2024.
\newblock URL \url{https://openreview.net/forum?id=KqbCvIFBY7}.

\bibitem[Esser et~al.(2024)Esser, Kulal, Blattmann, Entezari, M{\"{u}}ller, Saini, Levi, Lorenz, Sauer, Boesel, Podell, Dockhorn, English, and Rombach]{DBLP:conf/icml/EsserKBEMSLLSBP24}
Patrick Esser, Sumith Kulal, Andreas Blattmann, Rahim Entezari, Jonas M{\"{u}}ller, Harry Saini, Yam Levi, Dominik Lorenz, Axel Sauer, Frederic Boesel, Dustin Podell, Tim Dockhorn, Zion English, and Robin Rombach.
\newblock Scaling rectified flow transformers for high-resolution image synthesis.
\newblock In \emph{Forty-first International Conference on Machine Learning, {ICML} 2024, Vienna, Austria, July 21-27, 2024}. OpenReview.net, 2024.
\newblock URL \url{https://openreview.net/forum?id=FPnUhsQJ5B}.

\bibitem[Favero et~al.(2025)Favero, Saremi, Kaczmarek, Nichyporuk, and Arbel]{DBLP:journals/corr/abs-2502-03687}
Gian~Mario Favero, Parham Saremi, Emily Kaczmarek, Brennan Nichyporuk, and Tal Arbel.
\newblock Conditional diffusion models are medical image classifiers that provide explainability and uncertainty for free.
\newblock \emph{CoRR}, abs/2502.03687, 2025.
\newblock \doi{10.48550/ARXIV.2502.03687}.
\newblock URL \url{https://doi.org/10.48550/arXiv.2502.03687}.

\bibitem[Hjort and Jones(1996)]{HjortJones1996}
Nils~Lid Hjort and M.~C. Jones.
\newblock Locally parametric nonparametric density estimation.
\newblock \emph{Annals of Statistics}, 24\penalty0 (4):\penalty0 1619--1647, 1996.
\newblock \doi{10.1214/aos/1032298288}.

\bibitem[Ho and Salimans(2022)]{DBLP:journals/corr/abs-2207-12598}
Jonathan Ho and Tim Salimans.
\newblock Classifier-free diffusion guidance.
\newblock \emph{CoRR}, abs/2207.12598, 2022.
\newblock \doi{10.48550/ARXIV.2207.12598}.
\newblock URL \url{https://doi.org/10.48550/arXiv.2207.12598}.

\bibitem[Kendall(1945)]{kendall1945treatment}
Maurice~G Kendall.
\newblock The treatment of ties in ranking problems.
\newblock \emph{Biometrika}, 33\penalty0 (3):\penalty0 239--251, 1945.

\bibitem[Kish(1965)]{Kish1965}
Leslie Kish.
\newblock \emph{Survey Sampling}.
\newblock John Wiley \& Sons, New York, 1965.

\bibitem[Kulesza et~al.(2012)Kulesza, Taskar, et~al.]{kulesza2012determinantal}
Alex Kulesza, Ben Taskar, et~al.
\newblock Determinantal point processes for machine learning.
\newblock \emph{Foundations and Trends{\textregistered} in Machine Learning}, 5\penalty0 (2--3):\penalty0 123--286, 2012.

\bibitem[Kuznetsova et~al.(2020)Kuznetsova, Rom, Alldrin, Uijlings, Krasin, Pont{-}Tuset, Kamali, Popov, Malloci, Kolesnikov, Duerig, and Ferrari]{DBLP:journals/ijcv/KuznetsovaRAUKP20}
Alina Kuznetsova, Hassan Rom, Neil Alldrin, Jasper R.~R. Uijlings, Ivan Krasin, Jordi Pont{-}Tuset, Shahab Kamali, Stefan Popov, Matteo Malloci, Alexander Kolesnikov, Tom Duerig, and Vittorio Ferrari.
\newblock The open images dataset {V4}.
\newblock \emph{Int. J. Comput. Vis.}, 128\penalty0 (7):\penalty0 1956--1981, 2020.
\newblock \doi{10.1007/S11263-020-01316-Z}.
\newblock URL \url{https://doi.org/10.1007/s11263-020-01316-z}.

\bibitem[Labs(2024)]{flux2024}
Black~Forest Labs.
\newblock Flux.
\newblock \url{https://github.com/black-forest-labs/flux}, 2024.

\bibitem[Labs et~al.(2025)Labs, Batifol, Blattmann, Boesel, Consul, Diagne, Dockhorn, English, English, Esser, Kulal, Lacey, Levi, Li, Lorenz, Müller, Podell, Rombach, Saini, Sauer, and Smith]{labs2025flux1kontextflowmatching}
Black~Forest Labs, Stephen Batifol, Andreas Blattmann, Frederic Boesel, Saksham Consul, Cyril Diagne, Tim Dockhorn, Jack English, Zion English, Patrick Esser, Sumith Kulal, Kyle Lacey, Yam Levi, Cheng Li, Dominik Lorenz, Jonas Müller, Dustin Podell, Robin Rombach, Harry Saini, Axel Sauer, and Luke Smith.
\newblock Flux.1 kontext: Flow matching for in-context image generation and editing in latent space, 2025.
\newblock URL \url{https://arxiv.org/abs/2506.15742}.

\bibitem[Lin(1991)]{DBLP:journals/tit/Lin91}
Jianhua Lin.
\newblock Divergence measures based on the shannon entropy.
\newblock \emph{{IEEE} Trans. Inf. Theory}, 37\penalty0 (1):\penalty0 145--151, 1991.
\newblock \doi{10.1109/18.61115}.
\newblock URL \url{https://doi.org/10.1109/18.61115}.

\bibitem[Lipman et~al.(2023)Lipman, Chen, Ben{-}Hamu, Nickel, and Le]{DBLP:conf/iclr/LipmanCBNL23}
Yaron Lipman, Ricky T.~Q. Chen, Heli Ben{-}Hamu, Maximilian Nickel, and Matthew Le.
\newblock Flow matching for generative modeling.
\newblock In \emph{The Eleventh International Conference on Learning Representations, {ICLR} 2023, Kigali, Rwanda, May 1-5, 2023}. OpenReview.net, 2023.
\newblock URL \url{https://openreview.net/forum?id=PqvMRDCJT9t}.

\bibitem[Liu et~al.(2023)Liu, Gong, and Liu]{DBLP:conf/iclr/LiuG023}
Xingchao Liu, Chengyue Gong, and Qiang Liu.
\newblock Flow straight and fast: Learning to generate and transfer data with rectified flow.
\newblock In \emph{The Eleventh International Conference on Learning Representations, {ICLR} 2023, Kigali, Rwanda, May 1-5, 2023}. OpenReview.net, 2023.
\newblock URL \url{https://openreview.net/forum?id=XVjTT1nw5z}.

\bibitem[Liu et~al.(2026)Liu, Li, and Nguyen]{liu2026consistency}
Xinshuang Liu, Runfa~Blark Li, and Truong Nguyen.
\newblock Consistency-preserving diverse video generation.
\newblock \emph{arXiv preprint arXiv:2602.15287}, 2026.

\bibitem[Loader(1996)]{Loader1996}
Clive~R. Loader.
\newblock Local likelihood density estimation.
\newblock \emph{Annals of Statistics}, 24\penalty0 (4):\penalty0 1602--1618, 1996.
\newblock \doi{10.1214/aos/1032298287}.

\bibitem[Loftsgaarden and Quesenberry(1965)]{loftsgaarden1965nonparametric}
Don~O Loftsgaarden and Charles~P Quesenberry.
\newblock A nonparametric estimate of a multivariate density function.
\newblock \emph{The Annals of Mathematical Statistics}, 36\penalty0 (3):\penalty0 1049--1051, 1965.

\bibitem[Loshchilov and Hutter(2019)]{DBLP:conf/iclr/LoshchilovH19}
Ilya Loshchilov and Frank Hutter.
\newblock Decoupled weight decay regularization.
\newblock In \emph{7th International Conference on Learning Representations, {ICLR} 2019, New Orleans, LA, USA, May 6-9, 2019}. OpenReview.net, 2019.
\newblock URL \url{https://openreview.net/forum?id=Bkg6RiCqY7}.

\bibitem[Mason and Handscomb(2002)]{mason2002chebyshev}
John~C Mason and David~C Handscomb.
\newblock \emph{Chebyshev polynomials}.
\newblock Chapman and Hall/CRC, 2002.

\bibitem[Moffat et~al.(2022)Moffat, Mackenzie, Thomas, and Azzopardi]{DBLP:conf/sigir/MoffatM0A22}
Alistair Moffat, Joel Mackenzie, Paul Thomas, and Leif Azzopardi.
\newblock A flexible framework for offline effectiveness metrics.
\newblock In Enrique Amig{\'{o}}, Pablo Castells, Julio Gonzalo, Ben Carterette, J.~Shane Culpepper, and Gabriella Kazai, editors, \emph{{SIGIR} '22: The 45th International {ACM} {SIGIR} Conference on Research and Development in Information Retrieval, Madrid, Spain, July 11 - 15, 2022}, pages 578--587. {ACM}, 2022.
\newblock \doi{10.1145/3477495.3531924}.
\newblock URL \url{https://doi.org/10.1145/3477495.3531924}.

\bibitem[Morshed and Boddeti(2025)]{DBLP:conf/cvpr/MorshedB25}
Mashrur~M. Morshed and Vishnu Boddeti.
\newblock Diverseflow: Sample-efficient diverse mode coverage in flows.
\newblock In \emph{{IEEE/CVF} Conference on Computer Vision and Pattern Recognition, {CVPR} 2025, Nashville, TN, USA, June 11-15, 2025}, pages 23303--23312. Computer Vision Foundation / {IEEE}, 2025.
\newblock \doi{10.1109/CVPR52734.2025.02170}.
\newblock URL \url{https://openaccess.thecvf.com/content/CVPR2025/html/Morshed\_DiverseFlow\_Sample-Efficient\_Diverse\_Mode\_Coverage\_in\_Flows\_CVPR\_2025\_paper.html}.

\bibitem[Perez et~al.(2018)Perez, Strub, De~Vries, Dumoulin, and Courville]{perez2018film}
Ethan Perez, Florian Strub, Harm De~Vries, Vincent Dumoulin, and Aaron Courville.
\newblock Film: Visual reasoning with a general conditioning layer.
\newblock In \emph{Proceedings of the AAAI conference on artificial intelligence}, volume~32, 2018.

\bibitem[Rombach et~al.(2022)Rombach, Blattmann, Lorenz, Esser, and Ommer]{Rombach_2022_CVPR}
Robin Rombach, Andreas Blattmann, Dominik Lorenz, Patrick Esser, and Bj\"orn Ommer.
\newblock High-resolution image synthesis with latent diffusion models.
\newblock In \emph{Proceedings of the IEEE/CVF Conference on Computer Vision and Pattern Recognition (CVPR)}, pages 10684--10695, June 2022.

\bibitem[Silverman(1986)]{DBLP:books/sp/Silverman86}
Bernard~W. Silverman.
\newblock \emph{Density Estimation for Statistics and Data Analysis}.
\newblock Springer, 1986.
\newblock ISBN 978-1-4899-3324-9.
\newblock \doi{10.1007/978-1-4899-3324-9}.
\newblock URL \url{https://doi.org/10.1007/978-1-4899-3324-9}.

\bibitem[Song et~al.(2021)Song, Sohl{-}Dickstein, Kingma, Kumar, Ermon, and Poole]{DBLP:conf/iclr/0011SKKEP21}
Yang Song, Jascha Sohl{-}Dickstein, Diederik~P. Kingma, Abhishek Kumar, Stefano Ermon, and Ben Poole.
\newblock Score-based generative modeling through stochastic differential equations.
\newblock In \emph{9th International Conference on Learning Representations, {ICLR} 2021, Virtual Event, Austria, May 3-7, 2021}. OpenReview.net, 2021.
\newblock URL \url{https://openreview.net/forum?id=PxTIG12RRHS}.

\bibitem[Spearman(1987)]{spearman1987proof}
Charles Spearman.
\newblock The proof and measurement of association between two things.
\newblock \emph{The American journal of psychology}, 100\penalty0 (3/4):\penalty0 441--471, 1987.

\bibitem[Wald(1949)]{wald1949note}
Abraham Wald.
\newblock Note on the consistency of the maximum likelihood estimate.
\newblock \emph{The Annals of Mathematical Statistics}, 20\penalty0 (4):\penalty0 595--601, 1949.

\bibitem[Wand and Jones(1995)]{WandJones1995}
M.~P. Wand and M.~C. Jones.
\newblock \emph{Kernel Smoothing}, volume~60 of \emph{Monographs on Statistics and Applied Probability}.
\newblock Chapman and Hall/CRC, London, 1995.
\newblock \doi{10.1201/b14876}.

\bibitem[Yu et~al.(2023)Yu, Wang, Zhao, Ghanem, and Zhang]{yu2023freedom}
Jiwen Yu, Yinhuai Wang, Chen Zhao, Bernard Ghanem, and Jian Zhang.
\newblock Freedom: Training-free energy-guided conditional diffusion model.
\newblock In \emph{Proceedings of the IEEE/CVF International Conference on Computer Vision}, pages 23174--23184, 2023.

\bibitem[Zhang et~al.(2018)Zhang, Isola, Efros, Shechtman, and Wang]{zhang2018unreasonable}
Richard Zhang, Phillip Isola, Alexei~A Efros, Eli Shechtman, and Oliver Wang.
\newblock The unreasonable effectiveness of deep features as a perceptual metric.
\newblock In \emph{Proceedings of the IEEE conference on computer vision and pattern recognition}, pages 586--595, 2018.

\end{thebibliography}

\newpage
\onecolumn
\title{Score-Regularized Joint Sampling with Importance Weights for Flow Matching}
\maketitle
\appendix

\section{Importance Sampling}
\label{app:importance_sampling}

For completeness, we briefly introduce importance sampling here. While drawing non-IID samples can improve coverage of the support, the marginal density of each sample is generally biased. To estimate the expectation of a test function $f$ as
\begin{equation}
    \mathbb{E}_{X\sim p}[f(X)]\approx\frac{1}{n}\sum_{i=1}^n f\!\left(X^{(i)}\right), \quad X^{(i)} \overset{\text{IID}}{\sim} p,
\end{equation}
one requires IID samples from $p$. Under non-IID sampling, the samples are drawn from a joint distribution
\begin{equation}
(X^{(1)},\ldots,X^{(n)}) \sim p_\text{joint}.
\end{equation}
This can increase diversity but introduces bias that must be corrected with importance weights:
\begin{equation}
\hat\mu = \frac{1}{n}\sum_{i=1}^n w\!\left(X^{(i)}\right)\,f\!\left(X^{(i)}\right).
\end{equation}
We seek $w$ such that 
\begin{equation}
    \mathbb{E}_{(X^{(1)},\ldots,X^{(n)}) \sim p_\text{joint}} \hat\mu = \mathbb{E}_{X\sim p}[f(X)].
\end{equation}
It is straightforward to verify that 
\begin{equation}
w(x)=\frac{p(x)}{p'(x)},
\end{equation}
suffices, where $p'$ is the marginal density of $p_\text{joint}$,
\begin{equation}
p'_i(x)
= \int p_{\text{joint}}\big(x^{(1)},\ldots,x^{(n)}\big)\Big|_{x^{(i)} = x}\,\mathrm{d}x^{(-i)},
\end{equation}
and, assuming exchangeability,
\begin{equation}
p'(x) := p'_1(x)=\cdots= p'_n(x).
\end{equation}
\emph{Proof.} Substituting the proposed weights, we obtain
\begin{equation}
\hat\mu = \frac{1}{n}\sum_{i=1}^n w\!\left(X^{(i)}\right)\,f\!\left(X^{(i)}\right),
\quad
\,w(x)=\frac{p(x)}{p'(x)}.
\end{equation}
Thus,
\begin{equation}
\begin{aligned}
& \mathbb{E}_{(X^{(1)},\ldots,X^{(n)})\sim p_\text{joint}}[\hat\mu] \\
&= \int \frac{1}{n}\sum_{i=1}^n  f(x^{(i)})\,w(x^{(i)})\,p_\text{joint}\big(x^{(1:n)}\big)\,\mathrm{d}x^{(1:n)}\\
&= \frac{1}{n} \sum_{i=1}^n \int f(x^{(i)})\,w(x^{(i)})\,p'_i(x^{(i)})\,\mathrm{d}x^{(i)} \\
&= \frac{1}{n} \sum_{i=1}^n \int f(x^{(i)})\,p(x^{(i)})\,\mathrm{d}x^{(i)} \\
&= \mathbb{E}_{X\sim p}\big[f(X)\big].
\end{aligned}
\end{equation}

\section{Relationship Between the Score and the Velocity in Rectified Flow}
\label{app:score_function}

For completeness, we include the known relationship between the score and the velocity in rectified flow here.

\paragraph{Setup.}
Let $X_0\sim p_0=\mathcal N(0,I)$ and $X_1\sim p_{\text{data}}$ independent. Given $X_1$ and $X_0$, $X_t=(1-t)X_0+tX_1$ with $t\in(0,1)$. Denote by $p_t$ the density of $X_t$ and by $s(x,t):=\nabla_x\log p_t(x)$ its score. 

\paragraph{Marginal density of $X_t$.}
Solving for $X_0$ gives
\begin{equation}
X_0=\tfrac{X_t-tX_1}{1-t}.
\end{equation}
For fixed $X_1$, the map $x_0\mapsto (1-t)x_0+tX_1$ has Jacobian $(1-t)I$ and determinant $(1-t)^d$. By change of variables,
\begin{equation}
p_{X_t\mid X_1}(x\mid X_1)=(1-t)^{-d}\,p_0\!\Big(\tfrac{x-tX_1}{1-t}\Big),
\end{equation}
hence
\begin{equation}
p_t(x)=\E_{X_1}\!\left[(1-t)^{-d}p_0\!\Big(\tfrac{x-tX_1}{1-t}\Big)\right].
\label{eq:pt_mixture}
\end{equation}

\paragraph{Score as a conditional mean.}
Differentiating Eq.~(\ref{eq:pt_mixture}) and using $\nabla\log p_0(z)=-z$ for $\mathcal N(0,I)$ gives
\begin{equation}
\begin{aligned}
& \quad \nabla_x \log p_t(x) \\
&=\frac{\E_{X_1}\!\left[(1-t)^{-d}p_0(\cdot)\,\nabla_x\log p_0\!\big(\tfrac{x-tX_1}{1-t}\big)\right]}{\E_{X_1}\!\left[(1-t)^{-d}p_0(\cdot)\right]} \\
&= -\frac{1}{1-t}\,
\frac{\E_{X_1}\!\left[(1-t)^{-d}p_0(\cdot)\,\tfrac{x-tX_1}{1-t}\right]}{\E_{X_1}\!\left[(1-t)^{-d}p_0(\cdot)\right]} \\
&= -\frac{1}{1-t}\,\E\!\left[X_0\,\big|\,X_t=x\right].
\end{aligned}
\end{equation}
Thus
\begin{equation}
\E[X_0\mid X_t=x]=-(1-t)\,s(x,t).
\label{eq:m0}
\end{equation}

\paragraph{Linear constraint on conditional means.}
Taking conditional expectations of $X_t=(1-t)X_0+tX_1$ given $X_t=x$,
\begin{equation}
x=(1-t)\,\E[X_0\mid X_t=x]+t\,\E[X_1\mid X_t=x].
\label{eq:lin_relation}
\end{equation}
Let 
\begin{equation}
m_0(x,t):=\E[X_0\mid X_t=x]
\end{equation}
and 
\begin{equation}
m_1(x,t):=\E[X_1\mid X_t=x].
\end{equation}
Using Eq.~(\ref{eq:m0}) in Eq.~(\ref{eq:lin_relation}) gives
\begin{equation}
t\,m_1(x,t)=x+(1-t)^2 s(x,t).
\label{eq:m1}
\end{equation}

\paragraph{Velocity identity and conclusion.}
The rectified-flow velocity field is the conditional mean displacement,
\begin{equation}
\begin{aligned}
v_\theta(x,t) & =\E[X_1-X_0\mid X_t=x]\\
& =m_1(x,t)-m_0(x,t) \\
& =m_1(x,t)+(1-t)s(x,t),
\label{eq:vel_def}
\end{aligned}
\end{equation}
where the last equality uses Eq.~(\ref{eq:m0}). Multiplying Eq.~(\ref{eq:vel_def}) by $t$ and substituting Eq.~(\ref{eq:m1}) gives
\begin{equation}
t\,v_\theta(x,t)=x+(1-t)s(x,t),
\end{equation}
which implies
\begin{equation}
s(x,t)=\frac{t\,v_\theta(x,t)-x}{1-t}.
\end{equation}

\section{Evolution of Importance Weight}
\label{app:evolution_proofs}

\subsection{Evolution at a Fixed Position}
\label{app:log_w_fixed}

Represent the distribution as the outcome $X_1$ at $t=1$ of an ODE 
\begin{equation}
\dot{X}_t = \tilde v(X_t,t), \quad X_0\sim p_0,
\end{equation} 
where $\tilde v$ is either the original flow-matching velocity $v$ or the marginal velocity under non-IID sampling, $v + r_\phi$. From the continuity equation
\begin{equation}
\partial_t \tilde p_t(x) 
=-\nabla_x \cdot \big(\tilde v(x,t) \tilde p_t(x) \big),
\end{equation}
we obtain 
\begin{equation}
\begin{aligned}
& \quad \partial_t \log \tilde p_t(x) \\
& = - \frac{1}{\tilde p_t(x)} \nabla_x \cdot \big(\tilde v(x,t) \tilde p_t(x) \big) \\
& =-\nabla_x \cdot \tilde v(x,t) - \tilde v(x,t) \cdot \frac{\nabla_x \tilde p_t(x)}{\tilde p_t(x)} \\
& = -\nabla_x \cdot \tilde v(x,t) - \tilde v(x,t) \cdot \nabla_x \log \tilde p_t(x) \\
& = -\nabla_x \cdot \tilde v(x,t) - \tilde v(x,t) \cdot \tilde s(x,t).
\end{aligned}
\end{equation} 
Thus, 
\begin{equation}
    \partial_t \log p_t(x) = -\nabla_x \cdot v(x,t) - v(x,t) \cdot s(x,t),
    \label{eq:continuity_original}
\end{equation}
and
\begin{equation}
\begin{aligned}
\partial_t \log p''_{\phi,t}(x) & = -\nabla_x \cdot \big(v(x,t) + r_\phi(x,t)\big) \\
& \quad - \big(v(x,t) + r_\phi(x,t)\big) \cdot s''_\phi(x,t),
\label{eq:continuity_marginal}
\end{aligned}
\end{equation}
where $p,s$ denote the original flow and $p''_\phi,s''_\phi$ denote the marginal under diversity. Subtracting Eq.~(\ref{eq:continuity_marginal}) from Eq.~(\ref{eq:continuity_original}) gives
\begin{equation}
\begin{aligned}
    \partial_t \log w_t(x) & = \partial_t \log p_t(x) - \partial_t \log p''_{\phi,t}(x) \\
    & = \nabla_x \cdot r_\phi(x,t) + s''_\phi(x,t) \cdot r_\phi(x,t) \\
    & \quad + v(x,t) \cdot \big(s''_\phi(x,t) - s(x,t)\big).
    \label{eq:app_partial_t_log_w}
\end{aligned}
\end{equation}
which matches Theorem~\ref{theorem:log_w_fixed}.

\subsection{Evolution Along a Trajectory}
\label{app:log_w_along}

As in Definition~\ref{def:joint_ode}, samples evolve according to
\begin{equation}
\dot X_t^{(i)} = v \big(X_t^{(i)}, t\big) + u\big(X^{(i)}_t,X^{(-i)}_t,t\big), \!\!\!\!\!\quad X_0^{(i)} \overset{\text{IID}}{\sim} p_0,
\end{equation}
where $i$ indexes trajectories, $X^{(-i)}_t$ excludes $X^{(i)}_t$, $v$ is the pre-trained flow, and $u$ is a diversity velocity. For the importance weight along trajectory $i$,
\begin{equation}
\begin{aligned}
\frac{\mathrm{d}}{\mathrm{dt}} \log w_{\phi,t}(X_t^{(i)}) & = \nabla_x \log w_{\phi,t}(X_t^{(i)}) \cdot \dot X_t^{(i)} \\
& \quad + \partial_t \log w_{\phi,t}(X_t^{(i)}),
\label{eq:app_d_log_w}
\end{aligned}
\end{equation}
where
\begin{equation}
\begin{aligned}
\nabla_x \log w_{\phi,t} (x) & = \nabla_x \log p_t(x) - \nabla_x \log p''_{\phi,t}(x) \\
& = s(x,t) - s_\phi''(x,t).
\label{eq:app_partial_x_log_w}
\end{aligned}
\end{equation}
Combining Eq.~(\ref{eq:app_d_log_w}), Eq.~(\ref{eq:app_partial_t_log_w}), Eq.~(\ref{eq:app_partial_x_log_w}) yields
\begin{equation}
\begin{aligned}
& \frac{\mathrm{d}}{\mathrm{dt}} \log w_{\phi,t}(X_t^{(i)}) \\
& = \big(s(X_t^{(i)},t) - s_\phi''(X_t^{(i)},t)\big) \cdot \big(v \big(X_t^{(i)}, t\big) \\
& \quad\! + u\big(X^{(i)}_t,X^{(-i)}_t,t\big)\big) + \nabla_x \cdot r_\phi(X_t^{(i)},t) \\
& \quad\! + s''_\phi(X_t^{(i)},t) \cdot r_\phi(X_t^{(i)},t) \\
& \quad\! + v(X_t^{(i)},t) \cdot \big(s''_\phi(X_t^{(i)},t) - s(X_t^{(i)},t)\big) \\
& = \nabla_x \cdot r_\phi(X_t^{(i)},t) + s''_\phi(X_t^{(i)},t) \cdot r_\phi(X_t^{(i)},t) \\
& \quad\! - \!\big(s_\phi''(X_t^{(i)},t) - s(X_t^{(i)},t)\big) \cdot u\big(X^{(i)}_t,X^{(-i)}_t,t\big),
\label{eq:app_logw_along}
\end{aligned}
\end{equation}
which matches Theorem~\ref{theorem:log_w_along}.

\subsection{Evolution Along a Trajectory for Rectified Flows}
\label{app:log_w_along_rf}

For rectified flows, Appendix~\ref{app:score_function} yields
\begin{equation}
s(x,t) = \frac{t v(x,t) - x}{1-t}, 
\label{eq:app_s_original}
\end{equation}
and 
\begin{equation}
s''_\phi(x,t) = \frac{t \big(v(x,t) + r_\phi(x,t)\big) - x}{1-t}.
\label{eq:app_s_marginal}
\end{equation}
Thus, 
\begin{equation}
s''_\phi(x,t) - s(x,t) = \frac{t}{1-t} r_\phi(x,t).
\label{eq:app_delta_s}
\end{equation}
Substituting Eq.~(\ref{eq:app_s_original}), Eq.~(\ref{eq:app_s_marginal}), and Eq.~(\ref{eq:app_delta_s}) into \ref{eq:app_logw_along} gives
\begin{equation}
\begin{aligned}
& \quad \frac{\mathrm{d}}{\mathrm{dt}} \log w_{\phi,t}(X_t^{(i)}) \\
& = \nabla_x \cdot r_\phi(X_t^{(i)},t) \\
& \quad + \frac{t \big(v(X_t^{(i)},t) + r_\phi(X_t^{(i)},t)\big) - X_t^{(i)}}{1-t} \cdot r_\phi(X_t^{(i)},t) \\
& \quad - \frac{t}{1-t} r_\phi(X_t^{(i)},t) \cdot u\big(X^{(i)}_t,X^{(-i)}_t,t\big) \\
& = \nabla_x \cdot r_\phi(X_t^{(i)},t) + \frac{t v(X^{(i)}_t,t) - X^{(i)}_t}{1-t} \cdot r_\phi(X_t^{(i)},t) \\
& + \tfrac{t}{1-t} \big(r_\phi(X_t^{(i)},t) - u\big(X^{(i)}_t,X^{(-i)}_t,t\big)\big) \cdot r_\phi(X_t^{(i)},t).
\end{aligned}
\end{equation}
which matches Corollary~\ref{corollary:log_w_along_rf}.

\subsection{Evolution at a Fixed Position for Rectified Flows}
\label{app:log_w_fixed_rf}

Combining Eq.~(\ref{eq:continuity_original}) and Eq.~(\ref{eq:continuity_marginal}) with the rectified-flow identity in Appendix~\ref{app:score_function}, we obtain
\begin{equation}
\begin{aligned}
& \quad \partial_t \log w_t(x) \\
& = \nabla_x \cdot r_\phi(x,t) + v(x,t) \cdot \frac{t}{1-t} r_\phi(x,t)  \\
& \quad + \frac{t \big(v(x,t) + r_\phi(x,t)\big) - x}{1-t} \cdot r_\phi(x,t) \\
& = \nabla_x \cdot r_\phi(x,t) \\
& \quad + \frac{1}{1-t} r_\phi(x,t) \cdot \big(2 t v(x,t) + t r_\phi(x,t) - x\big).
\end{aligned}
\end{equation}
which matches Eq.~(\ref{eq:log_w_evolution_fixed}).

\section{Diversity Objectives Used in Experiments}

\label{app:diversity_velocities}

To ensure generality, we include several diversity objectives in our experiments.
\begin{itemize}
    \item \textbf{Determinantal Point Process (DPP)}~\citep{kulesza2012determinantal}: \\
    $h\big(x^{(1:n)}\big) = \log\det\!\big(\tfrac{\exp(-K)}{\exp(-K) + I}\big)$,
    \item \textbf{Harmonic DPP}: \\
    $h\big(x^{(1:n)}\big) = \log\det\!\big(\tfrac{K_\mathrm{H}}{K_\mathrm{H} + I}\big)$,
    \item \textbf{Particle Guidance (PG)}~\citep{DBLP:conf/iclr/CorsoXBBJ24}: \\
    $h\big(x^{(1:n)}\big) = -\sum_{i,j}\exp(-K_{ij})$,
    \item \textbf{Chebyshev}: \\
    $h\big(x^{(1:n)}\big) = -\sum_{r=1}^{R} \left[1 + \tfrac{2}{n}\sum_{i<j} T_r\!\big(\exp(-K_{ij})\big)\right]^2$,
    \item \textbf{Log-barrier}: \\
    $h\big(x^{(1:n)}\big) = \sum_{i<j} \log  \big(1 - \exp (-K_{ij})\big)$,
    \item \textbf{Reciprocal}: \\
    $h\big(x^{(1:n)}\big) = -\sum_{i<j} \frac{1}{K_{ij}}$,
\end{itemize}
where $K$ is defined in Eq.~(\ref{eq:K_definition}), $T_r$ is the $r$-th Chebyshev polynomial~\citep{mason2002chebyshev}, and $K_{\mathrm{H}}$ is defined as 
\begin{equation}
K_\mathrm{H} = \tfrac{1}{\sum_{r=0}^{n-1} a_r}\sum_{r=0}^{n-1} a_r\, T_r(L_{\cos}),
\end{equation}
with $a_0{=}1$, $a_r{=}2\!\left(1 - \tfrac{r}{n}\right)$ when $r{\ge}1$, $L_{\cos,ij} {=} \langle \hat x^{(i)}, \hat x^{(j)} \rangle$, and $\hat x^{(i)} {=} x^{(i)} / \|x^{(i)}\|$. Gradients are normalized in implementation, so constant scalings are omitted. For the strength parameter $\lambda$ of the diversity velocity (see Eq.~(\ref{eq:diversity_velocity})), we set $\lambda=1$ for Section~\ref{subsec:exp_gaussian_mixture}, $\lambda=0.1$ for Section~\ref{subsec:exp_text_to_image}, and $\lambda=0.3$ for Section~\ref{subsec:exp_inpainting}.

\section{Experiment Implementation Details}

\label{app:exp_details}

We first summarize the shared protocol for the two image-generation tasks in Table~\ref{tab:image_protocol}. Within each task, all methods use a \emph{matched} protocol: the same conditions, number of sampling steps, joint sample size $n$, number of repetitions, and IID reference size, with a common diversity strength $\lambda$ for the diversity-based methods, so that any difference in coverage radius is attributable to the methods themselves rather than to differences in the evaluation setup. Classifier-Free Guidance (CFG) is disabled for all image methods to isolate the effect of SR, and for inpainting the coverage radius is computed only over the masked regions.

\begin{table}[t]
\centering
\begin{tabular}{lllccccc}
\toprule
Task & Conditions & Steps & Sample size & Diversity strength & Repetitions & IID reference size \\
\midrule
Text-to-image & 6 text prompts & 28 & 10 & 0.1 & 1{,}000/prompt & 10{,}000/prompt \\
Inpainting & 10 masked images & 28 & 10 & 0.3 & 100/input & 1{,}000/input \\
\bottomrule
\end{tabular}
\caption{Consolidated protocol for the image-generation experiments.}
\label{tab:image_protocol}
\end{table}

For the Gaussian mixture experiment (Section~\ref{subsec:exp_gaussian_mixture}), we use 100 flow-matching steps with uniformly spaced time intervals and an Euler solver. For this task, we train the residual velocity for 100 epochs, sampling a random $t \in [0,1]$ for each example. We use the AdamW~\citep{DBLP:conf/iclr/LoshchilovH19} optimizer with a learning rate of $10^{-4}$, weight decay of $10^{-3}$, and a batch size of $1{,}000$.

For both text-to-image generation (Section~\ref{subsec:exp_text_to_image}) and image inpainting (Section~\ref{subsec:exp_inpainting}), we use 28 uniformly spaced sampling steps with an Euler solver and disable CFG~\citep{DBLP:journals/corr/abs-2207-12598} for all experiments, so that the measured differences isolate the effect of SR. The SR projection rule still applies with CFG enabled; one can use a proxy score direction read off from the CFG-guided velocity via Eq.~(\ref{eq:score_velocity}), which we leave for future work. We compute diversity objectives using the spatial mean of latent features. For inpainting, we further adopt a fixed text prompt, ``a realistic photo,'' across all input images.

\section{Parameterization and Architecture of Residual Velocity}
\label{app:rphi_arch}

The marginal ODE (Definition~\ref{def:marginal_ode}) requires the marginal velocity $v'$ of the non-IID, jointly drawn samples. Rather than learn $v'$ from scratch, we parameterize it as $v' = v + r_\phi$, reusing the pre-trained velocity $v$ and learning the correction $r_\phi$. When the diversity-velocity magnitude is small, the joint trajectories (Definition~\ref{def:joint_ode}) interact weakly, so $v'$ stays close to $v$ and the correction $r_\phi = v' - v$ is small; learning this small residual rather than a fresh velocity improves training efficiency and potentially enables a smaller network than $v$.

The residual velocity $r_\phi$ is a lightweight time-conditioned FiLM MLP with about $0.24$M parameters, used in the Gaussian-mixture importance-weight experiments (Section~\ref{subsec:exp_gaussian_mixture}; training details in Appendix~\ref{app:exp_details}). We detail its architecture and parameter count here for reproducibility.

\paragraph{Architecture.}
A sinusoidal embedding of $t$ is passed through a time MLP to a shared feature $e_t \in \mathbb{R}^{256}$, which conditions two FiLM blocks~\citep{perez2018film}, each with two width-$64$ linear layers, followed by a linear layer back to $\mathbb{R}^d$. For each layer, $e_t$ produces a FiLM scale $s$ and shift $b$ that modulate the activation as $h \leftarrow h \odot (1 + s_{\max}\tanh(s)) + b$ with $s_{\max}=1$, after which a SiLU nonlinearity is applied; the scale and shift projections are zero-initialized, so the conditioning starts from the identity. Figure~\ref{fig:rphi_arch} summarizes the architecture.

\begin{figure}[t]
\centering
\begin{tikzpicture}[font=\footnotesize,
  box/.style={draw, rounded corners, align=center, inner sep=4pt,
              minimum height=9mm, minimum width=16mm},
  io/.style={align=center}]
\node[io]  (t)    at (0,1.9)   {$t$};
\node[box] (sin)  at (2.0,1.9) {Sinusoidal\\embedding};
\node[box] (tmlp) at (4.8,1.9) {Time MLP\\$e_t\in\mathbb{R}^{256}$};
\node[io]  (xt)   at (0,0)    {$x_t\in\mathbb{R}^{d}$};
\node[box] (b1)   at (2.4,0)  {FiLM\\block 1};
\node[box] (b2)   at (5.0,0)  {FiLM\\block 2};
\node[box] (lin)  at (7.7,0)  {Linear\\$64\!\to\!d$};
\node[io]  (out)  at (10.1,0) {$r_\phi(x_t,t)$};
\draw[->] (t)   -- (sin);
\draw[->] (sin) -- (tmlp);
\draw[->] (xt)  -- (b1);
\draw[->] (b1)  -- (b2);
\draw[->] (b2)  -- (lin);
\draw[->] (lin) -- (out);
\draw[->,dashed] (tmlp.south) -- (b1.north);
\draw[->,dashed] (tmlp.south) -- (b2.north);
\node[font=\scriptsize] at (3.55,1.0) {FiLM scale/shift};
\end{tikzpicture}
\caption{\textbf{Architecture of the residual velocity $r_\phi$.} A sinusoidal time embedding feeds a time MLP that produces a shared feature $e_t\in\mathbb{R}^{256}$, which modulates both FiLM-conditioned blocks (dashed) through per-layer scale and shift. Each block contains two width-$64$ linear layers, each followed by FiLM and a SiLU activation; a final linear layer maps back to $\mathbb{R}^{d}$. Only the first input projection and the output layer depend on the data dimension $d$.}
\label{fig:rphi_arch}
\end{figure}

\paragraph{Parameter count.}
Table~\ref{tab:rphi_params} lists the per-module parameter count. Only the input projection and the output layer depend on the data dimension $d$, so the total, $242{,}944 + 129\,d$, grows linearly in $d$ and stays close to $0.24$M across the dimensions we use.

\begin{table}[t]
\centering
\caption{Trainable parameter count of the residual velocity $r_\phi$ given the data dimension $d$.}
\label{tab:rphi_params}
\begin{tabular}{lr}
\toprule
Module & Parameter count \\
\midrule
Time MLP & $98{,}816$ \\
Block 1 with FiLM & $64d + 70{,}016$ \\
Block 2 with FiLM & $74{,}112$ \\
Output layer & $65d$ \\
\midrule
\textbf{Total} & $\mathbf{242{,}944 + 129\,d}$ \\
\bottomrule
\end{tabular}
\end{table}

\section{Additional Experiments and Analyses}

\paragraph{Perceptual cross-check with LPIPS.}
To check that our latent coverage metric does not understate perceptual diversity, we recompute it under a perceptual LPIPS distance~\citep{zhang2018unreasonable} (LPIPS-Alex; Table~\ref{tab:lpips_crosscheck}); the two broadly agree at the IID-versus-diverse level.

\begin{table}[t]
\centering
\caption{\textbf{Perceptual cross-check of the text-to-image coverage radius (lower is better).} The ``Latent'' column is the model-free latent coverage radius (Mean column of Table~\ref{tab:text2image}); the ``LPIPS-Alex'' column recomputes it (Eq.~(\ref{eq:representation_error})) on the same samples under an AlexNet LPIPS distance. Values are mean(uncertainty) with $95\%$ confidence intervals. SR (soft and hard) is our method.}
\label{tab:lpips_crosscheck}
\begin{tabular}{l|c|cc}
\toprule
Method & SR & Latent$\downarrow$ & LPIPS-Alex$\downarrow$ \\
\midrule
IID & - & 20.9(2) & 3.877(36) \\
\midrule
\multirow{3}{*}{DPP} & \no   & 17.5(2) & 3.829(36) \\
                     & soft  & 17.4(2) & 3.829(36) \\
                     & hard  & 17.2(2) & 3.829(36) \\
\midrule
\multirow{3}{*}{PG}  & \no   & 16.9(2) & 3.873(37) \\
                     & soft  & 16.8(2) & 3.872(37) \\
                     & hard  & 16.6(2) & 3.870(37) \\
\midrule
\multirow{3}{*}{DF}  & \no   & 17.4(2) & 3.832(36) \\
                     & soft  & 17.3(2) & 3.831(36) \\
                     & hard  & 17.2(2) & 3.829(36) \\
\bottomrule
\end{tabular}
\end{table}

\paragraph{Score-based regularization vs.\ scalar annealing.}
SR changes the \emph{direction} of the diversity velocity, not its magnitude. Could rescaling the magnitude schedule $\beta(t)$ (the $\sqrt{1-t}$ time factor of $\gamma$ in Eq.~(\ref{eq:diversity_velocity}); default $\beta(t)=\sqrt{1-t}$) match its performance instead? On the $8$-D Gaussian setup (Table~\ref{tab:exp_gaussian_diversity_quality}), no swept $\beta(t)$ shape reaches the DPP$+$SR-hard reference (Table~\ref{tab:sr_vs_schedule}). This is a rough comparison, as the sweep ($500$ repetitions) and the reference ($10{,}000$ trials) are separate runs; still, since rescaling alone cannot match SR, its benefit appears to come from the directional, score-aware correction, not magnitude tuning.

\begin{table}[t]
\centering
\setlength{\tabcolsep}{4pt}
\begin{tabular}{l|ccc}
\toprule
Diversity velocity / schedule & Mode cov.\ $\uparrow$ & $\log p$ $\uparrow$ & RMSE $\downarrow$ \\
\midrule
DPP, no SR, $\beta=\sqrt{1-t}$ (default) & 8.53 & 7.077 & 0.529 \\
DPP, no SR, $\beta=$ cosine & 8.48 & 11.349 & 0.443 \\
DPP, no SR, $\beta=\max(1-t,0)$ (linear) & 8.44 & 15.925 & 0.325 \\
DPP, no SR, $\beta=$ sigmoidal & 8.33 & 17.149 & 0.286 \\
\midrule
DPP$+$SR-hard (ours) & 8.57 & 19.270 & 0.253 \\
\bottomrule
\end{tabular}
\caption{\textbf{Score-based regularization vs.\ scalar annealing schedules.} The first four rows disable SR and vary only the scalar magnitude schedule $\beta(t)$; the last row is the DPP$+$SR-hard reference from Table~\ref{tab:exp_gaussian_diversity_quality}.}
\label{tab:sr_vs_schedule}
\end{table}

\paragraph{Sensitivity to the diversity strength $\lambda$.}
The strength $\lambda$ (Eq.~(\ref{eq:diversity_velocity})) governs the diversity--quality trade-off: too small gives little diversity, too large pushes samples off-manifold and lowers quality. We sweep it for DPP$+$SR-hard on Stable Diffusion 3.5 Medium~\citep{DBLP:conf/icml/EsserKBEMSLLSBP24} text-to-image over $T_1$--$T_6$, reporting the LPIPS-Alex radius against an IID reference (Table~\ref{tab:lambda_sensitivity}; a separate $30$-repetition run, so its values are not comparable to Table~\ref{tab:text2image}). The radius traces a broad U-shape: it stays at or below IID for $\lambda$ up to about $0.15$ (minimum near $0.075$) and degrades for large $\lambda$ (clearly by $0.3$), so moderate $\lambda$ is robust and the default $\lambda=0.1$ is a safe choice.

\begin{table}[t]
\centering
\setlength{\tabcolsep}{6pt}
\begin{tabular}{l|c|c|c}
\toprule
DPP$+$SR-hard (ours) & LPIPS-Alex $\downarrow$ & $\Delta$ vs.\ IID & Beats IID? \\
\midrule
IID (reference) & 3.902 & - & - \\
\midrule
$\lambda=0.01$ & 3.880 & $-0.022$ & \yes \\
$\lambda=0.025$ & 3.849 & $-0.053$ & \yes \\
$\lambda=0.05$ & 3.826 & $-0.076$ & \yes \\
$\lambda=0.075$ & \textbf{3.816} & $\mathbf{-0.086}$ & \yes \\
$\lambda=0.1$ (default) & 3.854 & $-0.048$ & \yes \\
$\lambda=0.15$ & 3.851 & $-0.051$ & \yes \\
$\lambda=0.2$ & 3.923 & $+0.021$ & \no \\
$\lambda=0.3$ & 4.218 & $+0.316$ & \no \\
\bottomrule
\end{tabular}
\caption{\textbf{Sensitivity to the diversity strength $\lambda$} for DPP$+$SR-hard on Stable Diffusion 3.5 Medium text-to-image, averaged over $T_1$--$T_6$ (LPIPS-Alex radius, lower is better; $30$ repetitions, $B{=}10$). $\Delta$ is the gap to the IID reference and ``Beats IID?'' its sign at point estimates. The empirical minimum ($\lambda=0.075$) is in \textbf{bold}; the default $\lambda=0.1$ is marked.}
\label{tab:lambda_sensitivity}
\end{table}

\paragraph{Choice of baselines and related work.}
We compare against DPP, Particle Guidance (PG)~\citep{DBLP:conf/iclr/CorsoXBBJ24}, and DiverseFlow (DF)~\citep{DBLP:conf/cvpr/MorshedB25}, each evaluated with and without our SR, under the same matched, fixed-budget protocol (Tables~\ref{tab:exp_gaussian_diversity_quality} and~\ref{tab:text2image}). A related line is training-free guidance such as FreeDoM~\citep{yu2023freedom}, which steers a frozen pre-trained diffusion model toward a target condition; it guides a single sample, whereas we couple $n$ samples for mutual diversity and recover importance weights. We do not treat methods that construct or train the generative flow itself as baselines, since they are not focused on multi-sample diversity.

\section{Construction of Ground-Truth Density}
\label{app:llde_details}

We obtain ground-truth densities for the non-IID marginal by repeatedly resampling from the non-IID dataset and pooling the draws into a single large set. Given a sample set $\mathcal{Y}=\{y_i\}_{i=1}^B$, $y_i \in \mathbb{R}^d$, we evaluate densities at each sample $x \in \mathcal{Y}$ using a leave-one-out strategy---excluding $x$ from its own neighborhood---so that $n {=} B {-} 1$ samples contribute to the estimate. Letting $I_d$ denote the $d {\times} d$ identity, we follow prior work~\citep{Loader1996,HjortJones1996} and estimate $\log p'(x)$ by fitting a local quadratic model to $\log p'$ using kernel-weighted moments computed across multiple kNN-defined bandwidths (scales $h$). We then combine the resulting per-scale log-density estimates via inverse-variance weighting.

\paragraph{kNN neighborhoods and bandwidth selection.}
We use the neighbor-count grid
\begin{equation}
\mathcal{K} = \{10,15,25,35,50\}\cap\{1,\ldots,n\}.
\end{equation}
For each $k\in\mathcal{K}$, we sort all samples by their distance to $x$, after removing $x$ itself.  
Let $r_k(x)$ be the distance to the $k$-th nearest neighbor, and define the balloon bandwidth
\begin{equation}
h = \max\{r_k(x),\, h_{\min}\}, \quad h_{\min}=10^{-12}.
\end{equation}
To stabilize the moment estimates, we include more than $k$ neighbors:
\begin{equation}
K_{\mathrm{use}}(k) = \min\!\big\{\,n,\;\max(3k,\;k{+}8)\,\big\}.
\end{equation}
Let $\mathcal{N}^{(k)}_x$ be the indices of the $K_{\mathrm{use}}(k)$ nearest neighbors (with self-exclusion applied).

\paragraph{Kernel-weighted moments at scale $h$.}
Let the Gaussian \emph{base} kernel be $K(u)=(2\pi)^{-d/2}\exp(-\|u\|_2^2/2)$ and define the scaled kernel $K_h(u)=h^{-d}K(u/h)$, so that $K_h(u)=(2\pi)^{-d/2}h^{-d}\exp(-\|u\|_2^2/(2h^2))$. 
For offsets $u_i = y_i - x$ with $i\in\mathcal{N}^{(k)}_x$, set $w_i = K_h(u_i)$, and set $w_i{=}0$ for $i\notin\mathcal{N}^{(k)}_x$.  
Define the weighted moments (sums taken over $i{=}1,\ldots,n$)
\begin{equation}
m_0{=}\sum_i w_i,\quad\!\!\!\!
m_1{=}\sum_i w_i\,u_i,\quad\!\!\!\!
M_2{=}\sum_i w_i\,u_i u_i^\top,
\end{equation}
and the empirical mean and covariance
\begin{equation}
\mu = \frac{m_1}{m_0},\quad
S = \frac{M_2}{m_0} - \mu\mu^\top.
\end{equation}

\paragraph{Local log-density at scale $h$.}
Under the quadratic approximation
$\log p'(x{+}u) \approx a {+} b^\top u {+} \tfrac12 u^\top C u$,  
profiling the local likelihood yields the closed-form estimate
\begin{equation}
\begin{aligned}
\widehat a_h(x)
& = \log\!\Big(\frac{m_0}{n}\Big)
    + d\log h \\
& \quad - \tfrac12 \log\!\det S
    - \tfrac12\,\mu^\top S^{-1}\mu,
\end{aligned}
\end{equation}
and
\begin{equation}
\widehat p'_h(x) = \exp\{\widehat a_h(x)\},
\end{equation}
where $\widehat a_h(x)$ is the estimated local log-density at location $x$ and scale $h$.

\paragraph{Uncertainty at a single scale.}
For the \emph{unit-bandwidth} Gaussian kernel~\citep{WandJones1995}, the kernel roughness is
\begin{equation}
R(K) = \int K(u)^2\,\mathrm{d}u = (4\pi)^{-d/2}.
\end{equation}
Using Kish's effective sample size~\citep{Kish1965},
\begin{equation}
n_{\mathrm{eff}}(x,h)=\frac{(\sum_i w_i)^2}{\sum_i w_i^2},
\end{equation}
a Wald approximation~\citep{wald1949note} leads to
\begin{equation}
\Var(\widehat a_h(x)) \approx
\frac{R(K)}{\,n_{\mathrm{eff}}(x,h)\,h^{d}\,\widehat p'_h(x)},\\
\end{equation}
and
\begin{equation}
\begin{aligned}
I_h(x)
&= \Var(\widehat a_h(x))^{-1} \\
&= \frac{ n_{\mathrm{eff}}(x,h)\, h^d\, \widehat p'_h(x) }{ R(K) },
\end{aligned}
\end{equation}
where $I_h(x)$ is the per-scale information (inverse variance) associated with $\widehat a_h(x)$.

\paragraph{Multi-scale aggregation.}
We apply numerical checks at each scale $h$ and exclude any scale whose kernel weights are numerically unreliable. Covariance estimates are stabilized by symmetrization and a small ridge to enforce positive definiteness prior to inversion.

Given the valid per-scale estimates $\widehat a_h(x)$ and their information weights $I_h(x)$, we aggregate via inverse-variance weighting:
\begin{equation}
\widehat a(x)=\frac{\sum_{h} I_h(x)\,\widehat a_h(x)}{\sum_{h} I_h(x)},
\quad
\widehat p'(x)=\exp\{\widehat a(x)\}.
\end{equation}
Only scales that pass the numerical checks contribute to the final estimate.

\section{Baseline Density Estimators}
\label{app:baselines}

\paragraph{kNN density \citep{loftsgaarden1965nonparametric}.}
This method estimates $p'(x)$ by the $k$-NN ball volume. Let $r_k(x)$ be the distance to the $k$-th ($k{=}5$) nearest \emph{other} sample in $X=\{x^{(j)}\}_{j=1}^{B}\subset\mathbb{R}^d$, and let
$
V_d=\pi^{d/2}/\Gamma(\tfrac{d}{2}+1)
$
be the unit $d$-ball volume:
\begin{align}
\widehat{p}_{\text{kNN}}(x)
&=\;
\frac{k}{B\,V_d\,\big(r_k(x)\big)^d}.
\end{align}

\paragraph{Gaussian KDE with Silverman's rule \citep{DBLP:books/sp/Silverman86}.}
With isotropic bandwidth $h>0$,
\begin{align}
\widehat{p}_{\text{KDE}}(x)
&= \frac{1}{B}\sum_{j=1}^{B}
\frac{\exp\!\Big(-\frac{\|x-x^{(j)}\|_{2}^{2}}{2h^{2}}\Big)}{(2\pi)^{d/2}h^{d}}, \\ 
h
&= \sigma\left(\frac{4}{d+2}\right)^{\!\frac{1}{d+4}}\,
B^{-\frac{1}{d+4}}, \\
\sigma &= \tfrac{1}{d}\sum_{\ell=1}^{d} \text{std}(X_{:\ell}).
\end{align}

\paragraph{Multivariate Gaussian fit (MGF) \citep{DBLP:journals/jei/BishopN07}.}
This method fits $N(\mu,\Sigma)$ by maximum likelihood on all $B$ samples:
\begin{equation}
\begin{aligned}
\mu_{\text{ML}} &= \tfrac{1}{B}\sum_{j=1}^{B} x^{(j)}, \\
\Sigma_{\text{ML}} &= \tfrac{1}{B}\sum_{j=1}^{B} (x^{(j)}{-}\mu_{\text{ML}})(x^{(j)}{-}\mu_{\text{ML}})^\top.
\end{aligned}
\end{equation}
Evaluate the log-density at each $x^{(i)}$ under the same fitted model:
\begin{equation}
\begin{aligned}
& \quad \log \widehat{p}_{\text{MGF}}(x^{(i)}) \\
& = -\tfrac{1}{2}\!\Big(d\log(2\pi) + \log|\Sigma_{\text{ML}}|\Big) \\
& \quad -\tfrac{1}{2}\big(x^{(i)}{-}\mu_{\text{ML}}\big)^\top \Sigma_{\text{ML}}^{-1} \big(x^{(i)}{-}\mu_{\text{ML}}\big).
\end{aligned}
\end{equation}

\subsection{Downstream Expectation Estimation}
\label{app:expectation_full}

For the expectation $\mu{=}\mathbb{E} f(X)$ (Section~\ref{subsec:exp_gaussian_mixture}), evaluated by Jensen--Shannon (JS) divergence (Table~\ref{tab:js_results}), our trajectory-based weighting debiases the equal-weighted non-IID average and improves over additional IID samples. The density baselines reach lower JS, but they are given the analytic target density $p_1$ as their weight numerator (an oracle, available in closed form only on this example and never used by our estimator); the JS metric then rewards recovering $p_1$, even though their per-weight error and ranking are weaker than ours (Table~\ref{tab:exp_gaussian_mixture_density_comparison}).

\begin{table}[t]
\centering
\begin{tabular}{cc|ccc|cc}
\toprule
Ours & Ours$^{\dagger}$ & kNN & KDE & MGF & Equal & IID \\
\midrule
0.071(1) & 0.078(2) & 0.052(1) & 0.044(1) & 0.051(1) & 0.088(1) & 0.077(1) \\
\bottomrule
\end{tabular}
\caption{\textbf{Jensen--Shannon divergence for expectation estimation} (lower is better). Values are mean(uncertainty) with $95\%$ confidence intervals. $^{\dagger}$ is our fixed-position variant; ``Equal'' weights the non-IID samples equally and ``IID'' uses additional IID samples. The density-ratio estimators (kNN, KDE, MGF) are oracle-fed the analytic target density $p_1$ (see text).}
\label{tab:js_results}
\end{table}

\section{Metrics for Ranking}
\label{app:ranking_metrics}

\paragraph{Notation.}
Consider a query with $n$ items. Let $p_i$ denote the predicted score (the estimated importance weight in our experiment) and $g_i$ the ground-truth grade (the ground-truth importance weight in our experiment). 

\paragraph{Kendall's $\tau_b$ \citep{kendall1945treatment}.}
For every unordered pair $(i,j)$, define $s_p=\operatorname{sign}(p_i-p_j)$
and $s_g=\operatorname{sign}(g_i-g_j)$. 
Here, $s_p$ and $s_g$ summarize how the pair is ordered by the prediction and by the ground truth, respectively ($+1$ if the first item is larger, $-1$ if smaller, $0$ if tied).
Kendall's $\tau_b$ measures agreement between two orderings while accounting for ties. Count concordant pairs $C$ $(s_p=s_g\in\{\pm 1\})$, discordant pairs $D$ $(s_p=-s_g\in\{\pm 1\})$, and one-sided ties
$
T_p=\#\{(i,j): s_p=0,\,s_g\neq 0\}
$,
$
T_g=\#\{(i,j): s_g=0,\,s_p\neq 0\}
$.
Pairs tied in both lists ($s_p=s_g=0$) are ignored. Then
\begin{align}
\tau_b
&=\;
\frac{C - D}{
\sqrt{(C + D + T_p)\,(C + D + T_g)}
}\,, 
\end{align}
with $0$ as the result if the denominator is $0$. Values range from $-1$ (complete disagreement) to $+1$ (perfect agreement), with $0$ indicating no correlation.

\paragraph{Spearman's $\rho$ \citep{spearman1987proof}.}
Spearman's $\rho$ assesses the strength of a monotonic relationship between two rankings. Assign 1-based ranks (1 = best) with average-tied ranks and compute the Pearson correlation between predicted and ground-truth ranks:
\[
\rho \;=\; \mathrm{corr}\!\big(\mathbf{r}_{\text{pred}},\,\mathbf{r}_{\text{gt}}\big).
\]
Here, $\rho=1$ indicates identical rankings up to a monotonic transformation, $\rho=-1$ indicates perfect inverse order, and $\rho\approx 0$ indicates no monotonic association.

\paragraph{Graded Average Precision (gAP) \citep{DBLP:conf/sigir/MoffatM0A22}.}
Graded AP extends Average Precision to graded (multi-level) relevance. First, sort the data by predicted scores.
Normalize per query to $[0,1]$: shift if any grade is negative; then, if the maximum exceeds $1$, divide all grades by that maximum.
Define
\begin{align}
\mathrm{Prec}@i
&= \frac{\sum_{j=1}^{i} g_j}{i}, \quad
\mathrm{gAP} =
\frac{\sum_{i=1}^{n} g_i\,\mathrm{Prec}@i}{\sum_{i=1}^{n} g_i}. 
\end{align}
When $g_i\in\{0,1\}$, $\mathrm{gAP}$ reduces to standard AP. Larger values indicate that higher-graded items are concentrated near the top of the ranking.

\section{Additional Qualitative Samples}
\label{app:more_qualitative}

\begin{figure}[t]
    \centering
    \includegraphics[width=0.7\linewidth]{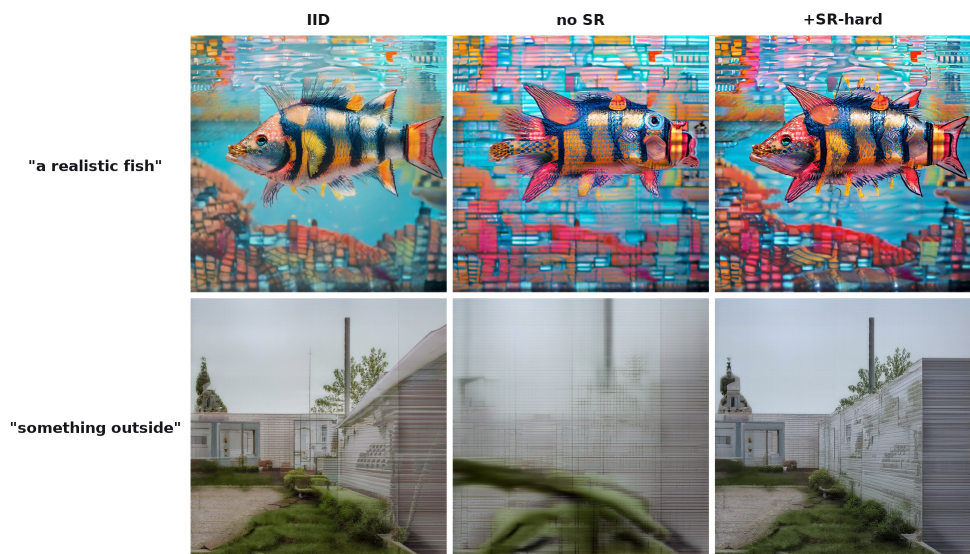}
    \caption{\textbf{Illustrative cases where SR changes the sample significantly.} Two selected text-to-image cases (base sampler: DPP) that illustrate SR's effect. Each row shows, at a matched seed, the IID reference, the base sampler without SR, and with SR-hard. In these cases SR recovers more plausible content while retaining the diversity the base sampler introduced.}
    \label{fig:app_sr_cases}
\end{figure}

Figure~\ref{fig:app_sr_cases} shows selected text-to-image cases where score-based regularization (SR) changes the sample significantly.

\begin{figure*}[t]
    \centering
    \includegraphics[width=0.7\linewidth]{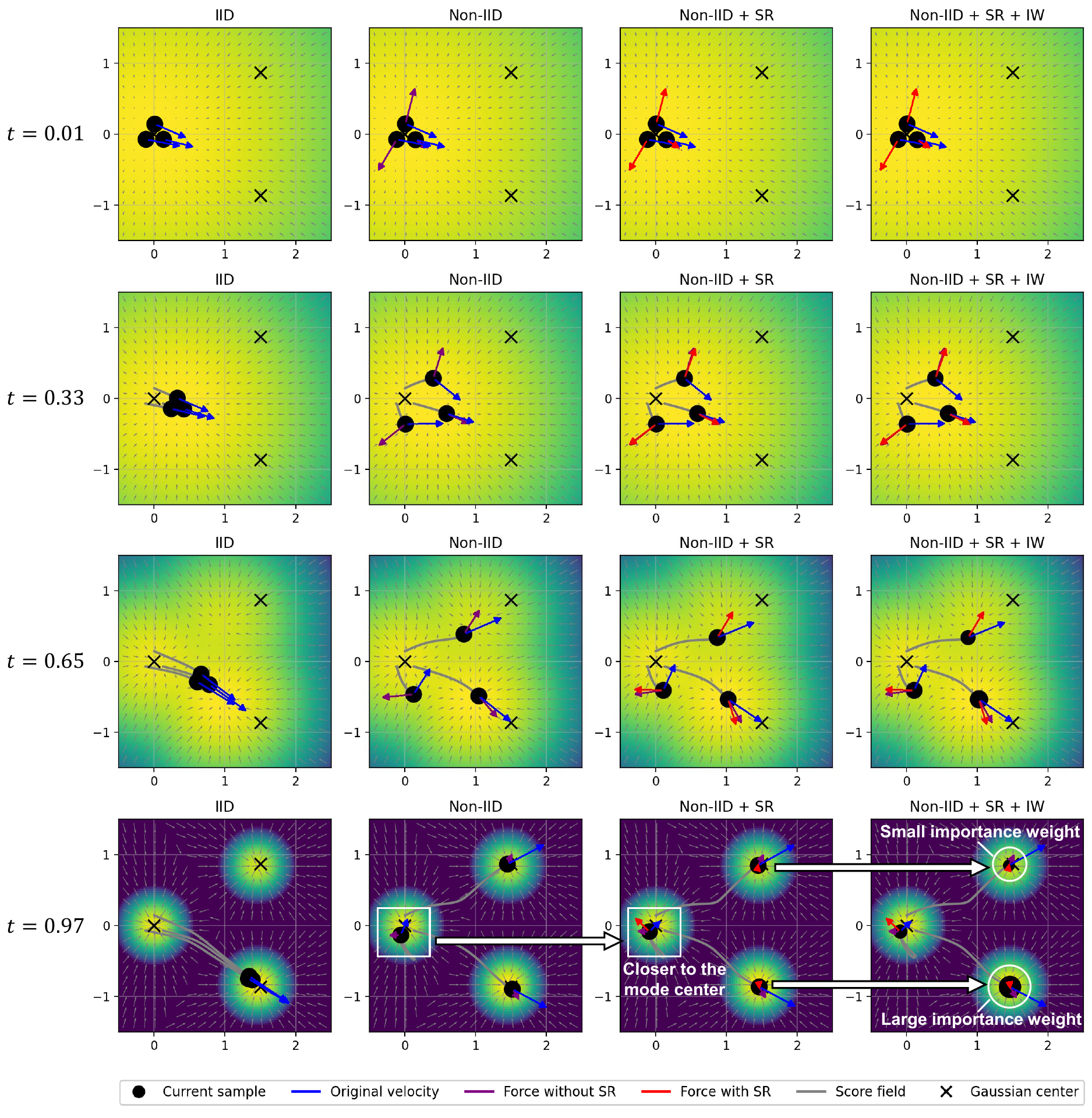}
    \caption{
    \textbf{Joint ODE trajectories on a three-component Gaussian mixture.}
    All methods start from identical initial states. ``IID'' is the baseline without a diversity objective and collapses to the highest-weight mode. ``Non-IID (DPP)'' uses a determinantal point process objective to encourage sample diversity, enabling the three trajectories to discover all three modes. ``SR'' denotes our score-based diversity-velocity regularization (the soft version is used in this example). Combining DPP$+$SR pulls samples toward the underlying modes while preserving coverage (white rectangle). ``IW'' denotes importance-weight estimation. The full method (DPP$+$SR$+$IW) further assigns an importance weight to each trajectory; larger markers indicate higher estimated weight (white circles). For visualization, arrow lengths are rescaled nonlinearly while preserving their relative ordering. The background shows the relative value of $\log p(x)$, where $p$ is the target density corresponding to the IID ODE; yellow indicates higher density and purple indicates lower density.
    }
    \label{fig:demo}
\end{figure*}

\section{Illustrative Example of Joint ODE Trajectories}

Figure~\ref{fig:demo} visualizes joint ODE trajectories on a 2D Gaussian mixture with three components of mixture weights $1/6$, $1/3$, and $1/2$. Under the IID baseline (no diversity objective), trajectories collapse to the highest-weight component, as expected. Introducing a DPP diversity objective yields \emph{Non-IID} trajectories that collectively cover all three modes, improving coverage under a fixed trajectory budget. Augmenting DPP with our score-based diversity-velocity regularization (SR) draws trajectories closer to the mode centers while preserving coverage (see the white rectangle). Finally, importance weighting (IW) assigns a weight to each trajectory. In the full method (DPP$+$SR$+$IW), the sample closest to the highest-weight component receives the largest weight (white circle) and that closest to the lowest-weight component receives the smallest, correcting the Non-IID sampling bias and yielding an unbiased estimator with respect to the IID target.

\end{document}